\pdfoutput=1
\documentclass[11pt]{article}

\usepackage{ACL2023}

\usepackage{times}
\usepackage{latexsym}
\usepackage{amsmath}
\usepackage{amsfonts}
\usepackage{multirow}
\usepackage{booktabs,arydshln}
\usepackage{graphicx}
\usepackage{algorithm}
\usepackage{algorithmic}
\usepackage{tabu}
\usepackage{xcolor}
\usepackage{hyperref}

\definecolor{myblue}{RGB}{99, 178, 238}

\makeatletter
\def\adl@drawiv#1#2#3{%
        \hskip.5\tabcolsep
        \xleaders#3{#2.5\@tempdimb #1{1}#2.5\@tempdimb}%
                #2\z@ plus1fil minus1fil\relax
        \hskip.5\tabcolsep}
\newcommand{\cdashlinelr}[1]{%
  \noalign{\vskip\aboverulesep
           \global\let\@dashdrawstore\adl@draw
           \global\let\adl@draw\adl@drawiv}
  \cdashline{#1}
  \noalign{\global\let\adl@draw\@dashdrawstore
           \vskip\belowrulesep}}
\makeatother

\newcommand{\rom}[1]{\uppercase\expandafter{\romannumeral #1\relax}}

\usepackage[T1]{fontenc}
\usepackage[utf8]{inputenc}

\usepackage{microtype}

\usepackage{inconsolata}

\title{Interpreting and Exploiting Functional Specialization in Multi-Head Attention under Multi-task Learning}

\author{Chong Li, Shaonan Wang, Yunhao Zhang, Jiajun Zhang, Chengqing Zong \\
        State Key Laboratory of Multimodal Artificial Intelligence Systems, \\
        Institute of Automation, CAS, Beijing, China\\
        School of Artificial Intelligence, University of Chinese Academy of Sciences, Beijing, China\\
        \{lichong2021, zhangyunhao2021\}@ia.ac.cn,\\
        \{shaonan.wang, jjzhang, cqzong\}@nlpr.ia.ac.cn
        }

\begin{document}
\maketitle
\begin{abstract}
%% Abstract 1
% Multi-head attention has been successfully applied to a variety of real-world tasks and even achieved super-human performance on several downstream tasks. 
% However, it is still unclear what mechanisms it has learned.
% The human brain is functionally specialized to efficiently handle multiple tasks. 
%% Abstract 2
% Existing neuroscience studies have found evidence of specialized brain regions tuned for functions from object recognition to navigation. 
% In contrast, Transformer-based models, even though achieve super-human performance on several downstream tasks, are regarded as a black box and utilized as a whole.
%% Abstract 3
Transformer-based models, even though achieving super-human performance on several downstream tasks, are often regarded as a black box and used as a whole.
It is still unclear what mechanisms they have learned, especially their core module: multi-head attention. 
Inspired by functional specialization in the human brain, which helps to efficiently handle multiple tasks, this work attempts to figure out whether the multi-head attention module will evolve similar function separation under multi-tasking training. 
If it is, can this mechanism further improve the model performance?
To investigate these questions, we introduce an interpreting method to quantify the degree of functional specialization in multi-head attention. 
We further propose a simple multi-task training method to increase functional specialization and mitigate negative information transfer in multi-task learning.
Experimental results on seven pre-trained transformer models have demonstrated that multi-head attention does evolve functional specialization phenomenon after multi-task training which is affected by the similarity of tasks. 
Moreover, the multi-task training strategy based on functional specialization boosts performance in both multi-task learning and transfer learning without adding any parameters. \footnote{Our code is available at \href{https://github.com/ZNLP/FunctionalSpecializationInMHA}{https://github.com/ZNLP/\\FunctionalSpecializationInMHA}}

\end{abstract}

\section{Introduction}
Transformer, based on the multi-head attention module, has been the dominant model for downstream applications due to its impressive results \citep{devlin-etal-2019-bert, brown-et-al-2020-gpt3, alexey2021vit}. 
However, it is still being utilized as a whole black-box model, and little is known about the functions of each sub-module on the final prediction. 
Simultaneously, although controversy still exists, there is overwhelming evidence that supports the idea of functional specialization in the human brain \citep{finger2001origins, kanwisher2010fucntioal}.
% Since the idea that the human brain consists of highly specialized components was proposed by Franz Joseph Gall (Viennese physician, 1758–1828), there have been overwhelming pieces of evidence, especially in the vision domain. 
Such a functional specialization mechanism makes it easier for the human brain to handle multiple tasks and solve new problems. 
It can reuse existing resources and at the same time evolve specific regions to avoid the huge cost of redesigning. 

% On the other hand, Transformer, as the dominant model for downstream applications, is still being utilized as a whole black-box model, and little is known about the functions of each sub-module on the final prediction. 
Considering the benefits of functional specialization to human learning ability, it is interesting to explore whether a transformer model, especially its central module multi-head attention, would evolve a similar mechanism under multi-task training. 
If so, which factors will impact the degree of functional specialization in the multi-head attention module? 
And how to exploit this phenomenon to improve the generalization ability of Transformer-based models? 

To investigate these questions, we first propose a method, called Important Attention-head Pruning (IAP), to quantify the degree of functional specialization in the multi-head attention of Transformer-based models.  
IAP first calculates the importance scores of each attention head on different tasks, then prunes the top important heads for each task to determine their impact on task performance. 
We apply our method to five different tasks with seven pre-trained transformers. 
% Results show that the multi-head attention module has evolved double dissociation phenomena in all task pairs, which is a significant indicator of functional specialization. % double dissociation phenomenon needs to explain.
Results show that the multi-head attention module has evolved distinct functional specialization phenomena across different sizes of BERT and pre-training methods. 
Further quantitative analysis indicates that there is a negative correlation between task similarity and the functional specialization phenomenon. 

Moreover, we propose a multi-task learning method, namely Important Attention-head Training (IAT), to promote the segregation of functions in the multi-head attention module by training only the most important part of attention heads for each task. 
Experimental results on the GLUE dataset have demonstrated that our method alleviates the negative transfer among tasks and improves the performance of Transformer-based models on both multi-task learning and transfer learning without additional parameters. 

% 不用说比单任务好，因为没有对比
% The multi-task learning models of four pre-trained Transformers, including $\text{BERT}_{\text{BASE}}$ and $\text{BERT}_{\text{LARGE}}$, even surpass corresponding single task fine-tuning models, which need nine times more parameters for all nine tasks.

To summarize, our main contributions are twofold:
\begin{itemize}
  \item We propose an interpretation method called IAP and find that the functional specialization phenomenon has evolved in multi-head attention after multi-task learning. Furthermore, empirical quantitative experiments show that such a phenomenon is influenced by the similarity between tasks: the more similar tasks are, the weaker the functional specialization phenomenon is.
  \item We propose an exploiting method called IAT to promote the degree of functional specialization. Experiments on multi-task learning and transfer learning validate that IAT is able to improve both the performance and generalization ability of multi-task learning models without adding any parameters.
  
\end{itemize}

\begin{figure*}[t]
\centering 
\setlength{\textfloatsep}{2pt}
\setlength{\intextsep}{2pt}
\setlength{\abovecaptionskip}{2pt}
\includegraphics[width=1\textwidth]{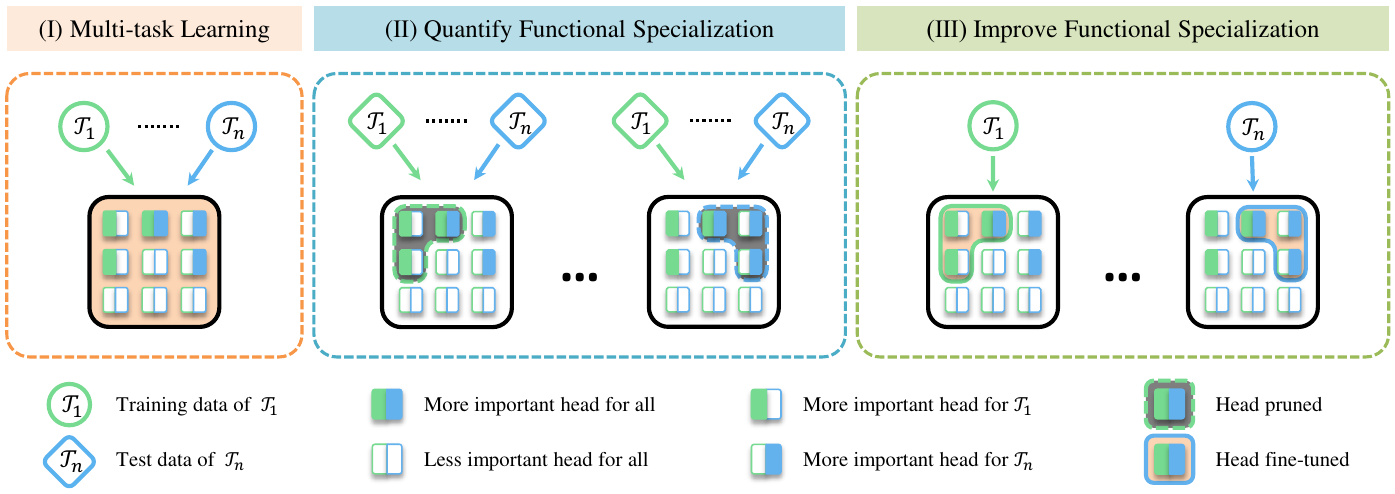}
\caption{Illustration of how to quantify and improve the degree of functional specialization in multi-head attention for Transformer-based models. Only attention heads, which are our research target, are depicted in the model for simplicity. (\rom{1}) Multi-task learning using Transformer-based models.  (\rom{2}) Quantify the functional specialization phenomenon by determining and pruning the important heads for each task. (\rom{3}) Improve the functional specialization phenomenon by only fine-tuning the important heads for each task in the last part of multi-task learning process. 
% 添加多任务学习的介绍
}\label{fig:framework}
\vspace{-5mm}
\end{figure*}

\section{Related Work}

\subsection{Interpreting Neural Networks}

\paragraph{Interpreting attention module}
Analogous to visual attention, the distribution of attention weight over input is often used to interpret the final decision of attention-based model \citep{clark-etal-2019-bert,vig-belinkov-2019-analyzing}. 
Therefore, a lot of work has been done to study the interpretability of attention distribution \citep{jain-wallace-2019-attention, serrano-smith-2019-attention,jacovi-goldberg-2020-towards} or design better explanation methods \citep{Brunner2020OnId, kobayashi-etal-2020-attention, bai-et-al-2021-why, lu-etal-2021-attention, liu2022rethinking}.
% 不相关的研究工作可以去掉，提一句，有大量的工作在研究注意力分布的可解释性
% However, \citet{jain-wallace-2019-attention} argued that it cannot be directly considered as feature importance, which means the distribution of attention weight does not provide a faithful explanation \citep{serrano-smith-2019-attention,jacovi-goldberg-2020-towards}. 
% On the contrary, \citet{wiegreffe-pinter-2019-attention} suggested that ``attention is not not explanation'' by challenging assumptions underlying \citet{jain-wallace-2019-attention} and proposing four alternative tests. 
% Subsequent studies strive to find ways to make attention weight more faithful or design better explanation methods \citep{Brunner2020OnId, mohankumar-etal-2020-towards, kobayashi-etal-2020-attention,tutek-snajder-2020-staying,bai-et-al-2021-why,liu2022rethinking}.

Our work can be classified into another line of study: investigating the individual attention head in the multi-head attention module. 
\citet{voita-etal-2019-analyzing} argued that there are redundant heads in Transformer by pruning less important heads and analyzing the resulting performance, which is confirmed by \citet{Michel2019Are16HeadsBetter}. 
\citet{jo-myaeng-2020-roles} analyzed the linguistic properties of the sentence representations from attention heads by ten linguistic probing tasks. 
\citet{hao-et-al-2021-interpreting-information} only retained the important heads in BERT and constructed an attribution tree to interpret the information interactions inside Transformer. 

Through pruning attention heads, we study the role they play in different tasks, rather than show redundancy in the multi-head attention module \citep{Michel2019Are16HeadsBetter}.

\paragraph{Interpretation inspired by neuroscience}
% It can be deleted
With more understanding of the functional specialization of the human brain, researchers attempt to interpret deep learning models with brain activities in specialized regions \citep{wehbe-etal-2014-aligning,toneva2019interpret,zhuang2021ventral,shahab2021visual}.
For example, \citet{toneva2019interpret} studied the representations of NLP models across different layers by aligning with two groups of brain areas among the language network. 

Unlike the existing works, we investigate whether the brain-like functional specialization phenomenon occurs in NLP models, and how to exploit this phenomenon to improve models. 

\subsection{Mitigating Negative Information Transfer in Multi-task Learning}
By joint learning multiple tasks, the performance of a model on the target task can be boosted with regularization or sharing parameters among tasks \citep{collobert2011multitask, ruder2017overview, liu-etal-2019-multi}.
However, multi-task learning models in NLP often suffer from negative information transfer and are inferior to the single task learning ones \citep{martinez-alonso-plank-2017-multitask,bingel-sogaard-2017-identifying}.

Our method aims to subdivide task-important modules in parameters shared to mitigate negative transfer among tasks, which is different from previous sampling or additional task-specific adapter methods \citep{Wu2020Understanding, pilault2021conditionally}. 
We only need to preserve mask variables for each attention head rather than all parameters during training \citep{Sun-et-al-2020-SparseSharing, lin-etal-2021-learning, xie-etal-2021-importance, liang-etal-2021-super}, which significantly reduce memory costs.

\section{Background}
% 为什么要这两块background，与研究问题有什么关系
\subsection{Multi-Head Attention Module}
Transformer \citep{Vaswani2017Attention} extended single head attention function to Multi-Head Attention (MHA) module, which aims at capturing information from different representation subspaces in parallel. 
Given input $X\in \mathbb{R}^{n\times d}$, this module linearly transforms it into $n_h$ subspaces and then applies attention separately:
\vspace{-1mm}
\begin{equation}
    \label{eq:attention}
    \vspace{-1mm}
    \small
    \begin{split}
        A_h(X) &= \text{Attention}(XW^{Q}_{h},XW^{K}_{h},XW^{V}_{h}) \\
        \text{with}&\ \text{Attention}(Q,K,V)=\text{softmax}(\frac{QK^T}{\sqrt{d_k}})V 
    \end{split}
\end{equation}
where $Q,K\in \mathbb{R}^{n\times d_k}$ and $V \in \mathbb{R}^{n\times d_v}$. The outputs of all heads are concatenated and linearly transformed into the output space of this module:
\vspace{-1mm}
\begin{equation}
    \vspace{-1.2mm}
    \small
    MHA(X)=[A_1(X);...;A_{n_h}(X)]W^{O}
% 	\vspace{-0.8mm}
\end{equation}
% Describe the multi-head self-attention network.

\subsection{Head Importance Score}
\label{sec:importance_score_method}
\citet{Michel2019Are16HeadsBetter} proposed an effective method to prune attention heads and evaluate the importance of attention heads for a task.
In order to prune the attention head $h$, they incorporated a mask variable $\xi_h \in [0,1]$ into the attention function:
\vspace{-0.5mm}
\begin{equation}
    \label{eq:mask_attention}
    \vspace{-0.5mm}
    \small
    \widetilde{A}_h(X) = \xi_{h}\cdot A_h(X)
    % \vspace{-0.5mm}
\end{equation}
and set it to a zero value.
When $\xi_h$ equals 1, Equation (\ref{eq:mask_attention}) is the same with the vanilla attention (Eq. (\ref{eq:attention})).
The head importance score $I_h^{(i)}$ of task $\mathcal{T}_i$ is approximated by the expected sensitivity of loss function to the mask variable $\xi_{h}$:
\vspace{-1mm}
\begin{equation}
    \vspace{-1mm}
    \small
    %I_h^{(i)}=\mathbb{E}_{x\sim X^{(i)}}\left| \frac{\partial{\mathcal{L}^{(i)}(x)}}{\partial{\xi_h}} \Bigr|_{\xi_h=1} \right|
    I_h^{(i)}=\mathbb{E}_{(x,y)\sim {\mathcal{D}}^{(i)}}\left| \frac{\partial{\mathcal{L}^{(i)}(x,y)}}{\partial{\xi_h}} \right|
    % \vspace{-1mm}
    \label{eq.importance_score}
\end{equation}
where ${\mathcal{D}}^{(i)}$ is the data distribution of task ${\mathcal{T}}_i$ and $\mathcal{L}^{(i)}(x,y)$ is the loss of task ${\mathcal{T}}_i$ on sample $(x,y)$.

Different from \citet{Michel2019Are16HeadsBetter} which prune the least important attention heads to prove the redundancy of attention heads, this paper focuses on exploring the functional specialization phenomenon after training, thus we prune the most important heads for each task. 

% \begin{figure*}[t]
% \centering 
% \setlength{\textfloatsep}{2pt}
% \setlength{\intextsep}{2pt}
% \setlength{\abovecaptionskip}{2pt}
% \includegraphics[width=1\textwidth]{imgs/d_score.png}
% \caption{Three different functional specialization cases in multi-head attention
% on dual-task learning ($\theta=30\%$). The first two-row images are views from distribution of important heads and relative performance after part of heads are pruned, respectively. (a) Both dissociation scores of two tasks are positive, which we called the double dissociation phenomenon. (b) There is a negative dissociation score between the two tasks. If the positive dissociation score is significantly high, e.g., task B, we can find a group of heads more important for this task. (c) The dissociation scores are both negative due to our wrong evaluation of the important head for each task.
% }\label{fig:dScoreCase}
% \vspace{-5mm}
% \end{figure*}

\section{Method}
Figure \ref{fig:framework} illustrates the general procedure of our methods. 
Firstly, Transformer-based models are utilized for multi-task learning and may arise segregation of functions in the multi-head attention module. 
Subsequently, the important attention heads are determined and pruned to quantify the functional specialization in multi-head attention (Section \ref{sec:interpret_method}). 
Lastly, the roles of important heads in each task are enhanced to promote the degree of functional specialization by important attention-head training (Section \ref{sec:IAT}).

\subsection{Interpreting: Important Attention-head Pruning}
\label{sec:interpret_method}
We introduce a two-step method, namely \textit{Important Attention-head Pruning} (IAP), to quantify the degree of functional specialization in multi-head attention.
First, the top $\alpha \in [0,1]$ percentage important heads $H^{\alpha}_i$ for task $\mathcal{T}_i$, e.g., the ones circled by dashed lines in \hyperref[fig:framework]{Figure 1(\rom{2})}, are found after dual-task or multi-task training by their head importance scores. 
Specifically, we calculate the head importance score $I_h^{(i)}$, defined by Eq. (\ref{eq.importance_score}), on training samples to approximate the contribution of head $h$ to task ${\mathcal{T}}_i$.

Second, dissociation experiments are conducted to determine the degree of functional specialization in multi-head attention.
Given a model $f_{\theta}$ after dual-task training on tasks ${\mathcal{T}}_A$ and ${\mathcal{T}}_B$, for example, the relative performance on ${\mathcal{T}}_A$ after pruning the top $\alpha$ important attention heads for ${\mathcal{T}}_B$, denoted by $H_{B}^{\alpha}$, is calculated as follows:
\vspace{-1mm}
\begin{equation}
    \vspace{-1mm}
    % \small
    \label{RP_2}
    \begin{array}{ccl}
    % RP_A(H_{B}^{\alpha}) &=& \frac{\textit{PERF}\left( f_{\theta \setminus H_{B}^{\alpha}}(X_A),Y_A\right)}{\textit{PERF}\left(f_{\theta}(X_A),Y_A\right)}  \\
    RP_A(H_{B}^{\alpha}) &=& \frac{\mathcal{P}\left( f_{\theta \setminus H_{B}^{\alpha}}(X_A),Y_A\right)}{\mathcal{P}\left(f_{\theta}(X_A),Y_A\right)}  \\
    \end{array}
\end{equation}
where $\mathcal{P}(\cdot)$ is the performance metric used, e.g., Accuracy, and $(X_A,Y_A)$ is the test sample of Task ${\mathcal{T}}_A$. 
Then, we estimate the degree of functional specialization by the relative performance difference after top $\alpha$ important heads for each task are pruned, called dissociation score:
\vspace{-0.5mm}
\begin{equation}
    \vspace{-0.5mm}
    % \small
    \label{DiffAndHar_2}
    \begin{array}{ccl}
    % D_{A}(\theta)       &=& RP_A(\theta, T_B) - RP_A(\theta, T_A),  \\
    % D_{B}(\theta)       &=& RP_B(\theta, T_A) - RP_B(\theta, T_B),  \\
    % D(\theta)           &=& \frac{D_A(\theta)+D_B(\theta)}{2}      \\
    D_{A}(\alpha)       &=& RP_A(H_{B}^{\alpha}) - RP_A(H_{A}^{\alpha}),  \\
    D_{B}(\alpha)       &=& RP_B(H_{A}^{\alpha}) - RP_B(H_{B}^{\alpha}),  \\
    D(\alpha)           &=& \frac{D_A(\alpha)+D_B(\alpha)}{2}      \\
    \end{array}
\end{equation}
where $D_{A}(\alpha)$ denotes the dissociation score of task ${\mathcal{T}}_A$, and $D(\alpha)$ is the average dissociation score of this dual-task learning.
Given an appropriate $\alpha$, a larger dissociation score implies a higher degree of functional specialization.

Similarly, the dissociation score of task $\mathcal{T}_i$ under multi-task learning is measured via:
% \vspace{-1mm}
\begin{equation}
    % \vspace{-1mm}
    % \small
    \label{DiffN}
    \begin{array}{rcl}
    % D_{i}(\theta)     &=& \frac{\sum_{j=1,j\neq i}^{n}RP_{i}(\theta, T_j)}{n-1} - RP_{i}(\theta, T_i),  \\
    % D(\theta) &=& \frac{\sum_{i=1}^{n}D_{i}(\theta)}{n}                          \\
    D_{i}(\alpha)     &=& \frac{\sum_{j=1,j\neq i}^{n}RP_{i}(H^{\alpha}_{j})}{n-1} - RP_{i}(H^{\alpha}_{i}),  \\
    D(\alpha) &=& \frac{\sum_{i=1}^{n}D_{i}(\alpha)}{n}                          \\
    \end{array}
\end{equation}

To clearly illustrate the functional specialization phenomenon, we summarize two representative cases under dual-task learning: 

\begin{itemize}
    \item \textbf{Double dissociation} when $D_A(\alpha)>0$ and $D_B(\alpha)>0$. 
    This is a significant indicator of functional specialization.
    That is, each task requires a unique group of heads, which can be selectively masked. 
    To eliminate the accidental functional specialization phenomenon, we argue that a distinct one occurs if the average dissociation score is higher than or equal to 10\%, i.e., $D(\alpha)\ge 10\%$, in which 10\% is chosen according to the definition of double dissociation in neuroscience \citep{shallice1988neuropsychology}.
    
    \item{\textbf{Single dissociation}} when $D_A(\alpha)>10\%$ and $D_B(\alpha)<0$, or $D_A(\alpha)<0$ and $D_B(\alpha)>10\%$. One significant positive dissociation score suggests functional specialization may only arise in this task. 

     % \item{\textbf{Others}} For example, $-5\%<D_A(\theta)<5\%$ and $-5\%<D_B(\theta)<5\%$. In this case, the dissociation scores of both tasks are relatively small, we argue that there is no functional specialization in the multi-head attention module. Specifically, the influence on all tasks will be almost unchanged when pruning different groups of heads.
\end{itemize}

    % The other cases, for example, $D_A(\alpha)=5\%$ and $D_B(\alpha)=-3\%$,  the dissociation scores of both tasks are relatively small. 
    The dissociation scores may be both negatives, which arise from the wrong evaluation of the important heads for each task. 
    It can be summarized into the double dissociation case under the correct evaluation and pruning. 
    In the other cases, e.g, the dissociation scores of both tasks are relatively small, 
    we argue that there is no functional specialization in the multi-head attention module. Specifically, the influence on all tasks will be almost identical when pruning different groups of heads.

\subsection{Exploiting: Important Attention-head Training}
\label{sec:IAT}
%Describe your multitask learning method here.
Motivated by the high degree of functional specialization in human brain, it is interesting to investigate whether a higher degree of functional specialization could improve the performance of the model on multi-task learning or transfer learning. 

To promote the degree of functional specialization in multi-head attention, we design a multi-task training method, named \textit{Important Attention-head Traning} (IAT). 
Specifically, only the top $\alpha \in [0,1]$ important attention heads for task $\mathcal{T}_i$ are tuned at the last $\delta \in [0,1]$ multi-task training process, and the parameters other than the multi-head attention module are trained as before. 
To achieve this, we introduce a mask variable $M_i\in \{0,1\}^{n_h}$ for task $\mathcal{T}_i$, where 1 indicates to fine-tune this attention head for $\mathcal{T}_i$. 
For example in \hyperref[fig:framework]{Figure 1(\rom{3})}, only the mask variables of heads circled by the solid blue line are set to 1 for $\mathcal{T}_n$. 
When $\alpha=1$ or $\delta=0$, our method is the same as the normal multi-task learning method. 

We expect to consolidate the roles of important heads for each task and facilitate the functional separation of multi-head attention in this way.

\section{Experimental Setup}
\subsection{Datasets}
We select a topic classification datasets \citep{zhang-et-al-2015-charCNN}, eight natural language understanding datasets of GLUE \citep{williams-etal-2018-broad,rajpurkar-etal-2016-squad,wang2018glue}, and two datasets \citep{maas-etal-2011-learning, khot-et-al-2018-scitail} for transfer learning in this study. 
% The WNLI dataset \citep{levesque-et-al-2012-wnli} in GLUE is not incorporated due to the small size of training samples, which is less than 1k.
To avoid an extreme ratio of training samples between tasks, only five large datasets in different tasks, which contain more than 10k training samples, are preserved in dual-task and multi-task learning interpretation experiments. 
Like \citet{karimi-mahabadi-etal-2021-parameter}, SciTail and IMDB are used only in transfer learning.
Statistics of all datasets used are shown in Table \ref{tab:Datasets}.
\setlength{\tabcolsep}{1mm}
\begin{table}[h]
	\centering

	\footnotesize
	% \vspace{-0.4cm}
	\renewcommand\arraystretch{1.1}
	\begin{center}
		\begin{tabular}{cccr}
			\toprule[1.2pt]  
			\textbf{Task}                        & \textbf{Dataset}   & \textbf{\#Class}   & \textbf{\#Train} \\
			\midrule[0.8pt]
   			Topic Classification        & AG's News${}^{\star}$ & 4 	    &   120,000 \\
			Acceptability               & CoLA      & 2         &   8,551  \\
                Natural Language Inference  & MNLI${}^{\star}$      & 3 	    &   392,702 \\
			Paraphrase                  & QQP${}^{\star}$       & 2 	    &   363,846 \\
                Paraphrase                  & MRPC      & 2 	    &   3,668   \\
			Question Answering          & QNLI${}^{\star}$      & 2 	    &   104,743 \\
			Sentiment Analysis          & SST-2${}^{\star}$     & 2 	    &   67,349 \\
                Entailment                  & RTE       & 2         &   2,490 \\
                Textual Similarity          & STS-B     & -         &   5,749 \\
                Natural Language Inference  & SciTail   & 2         &   23,596 \\
                % Natural Language Inference  & SNLI      & 3         &   549,367 \\
                Sentiment Analysis          & IMDB      & 2         &   25,000 \\
                % Object Classification       & CIFAR-10  & 10        &   50,000 \\
                % Joint Reasoning             & NLVR2     & 2         &   86,373 \\
			\bottomrule[1.2pt]
		\end{tabular}
		\caption{Statistic of datasets used. ${}^{\star}$ denotes dataset used in dual-task and multi-task interpretation experiments.}
		\label{tab:Datasets}
	\end{center}
	\vspace{-5mm}
\end{table}

\begin{figure*}[t]
\centering 
\setlength{\textfloatsep}{2pt}
\setlength{\intextsep}{2pt}
\setlength{\abovecaptionskip}{2pt}
\includegraphics[width=0.95\textwidth]{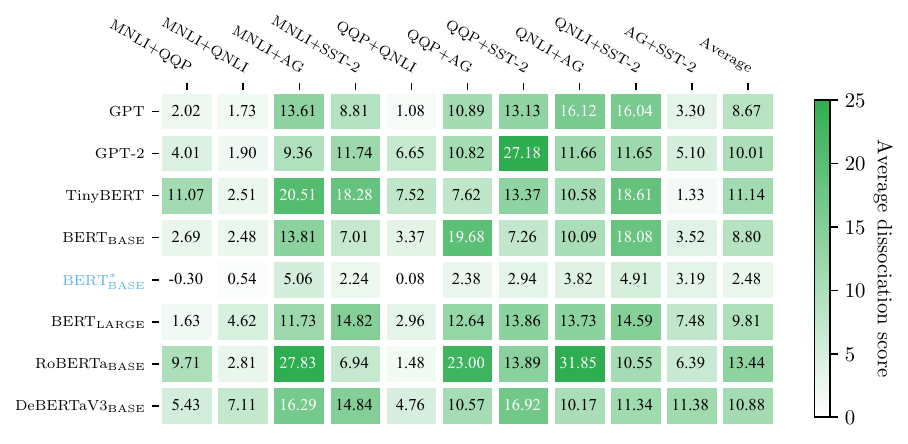}
\caption{Average dissociation scores of different Transformer-based models (y-axis) after ten dual-task learning tasks (x-axis) with $\alpha = 30\%$. The larger dissociation score implies a higher degree of functional specialization in multi-head attention (Section \ref{sec:interpret_method}). All dissociation scores are reported in Table \ref{tab:2tasks_all}. \textcolor{myblue}{${}^*$} indicates that the parameters of $\text{BERT}_{\text{BASE}}$ encoder are \textcolor{myblue}{\textbf{frozen}}, i.e., the output layers are fine-tuned only. }\label{fig:average_d_score}
\vspace{-4mm}
\end{figure*}

\subsection{Models}
% BERT Family 模型是常用的多任务学习模型，因此我们主要研究了这几个模型
As shown in Table \ref{tab:Models}, seven Pre-trained Transformer Models (PTMs), including GPT family models, different sizes of BERT and different pre-training methods \citep{radford2018gpt,radford2019gpt2,devlin-etal-2019-bert,liu2019roberta,jiao-etal-2020-tinybert,he2021debertav3}, are investigated in this paper. 
These models are all initialized from the transformer library of HuggingFace \cite{wolf2019huggingface}. 
Hyperparameters are reported in Appendix \ref{appendix:hyper_parameters}.
% \paragraph{Language Model}
% Tiny-BERT\cite{jiao-etal-2020-tinybert}, BERT-base-cased, BERT-Large-uncased\cite{devlin-etal-2019-bert}

% \paragraph{Vision-Language Model}
% ViLT\cite{kim2021vilt}, OFA\cite{wang2022ofa}.
\setlength{\tabcolsep}{1mm}
\begin{table}[h]
	\centering

	\footnotesize
	\vspace{-0.2cm}
	\renewcommand\arraystretch{0.9}
	\begin{center}
		\begin{tabular}{ccccc}
			\toprule[1.2pt]  
			\textbf{Model}                            & \textbf{\#L}       & \textbf{\#A} & \textbf{\#L} $\times$ \textbf{\#A} & \textbf{Parameters}  \\
			\midrule[0.8pt]
                GPT                              & 12        & 12  &   144 & 110M\\
                GPT-2                            & 24        & 16  &   384 & 355M\\
                TinyBERT                         & 6         & 12  &   72  & 67M\\
                $\text{BERT}_{\text{BASE}}$      & 12        & 12  &   144 & 110M\\ %102,267,648
                $\text{RoBERTa}_{\text{BASE}}$   & 12        & 12  &   144 & 125M\\ %124,646,146
                $\text{DeBERTaV3}_{\text{BASE}}$ & 12        & 12  &   144 & 184M\\ %183,832,064
                $\text{BERT}_{\text{LARGE}}$     & 24        & 16  &   384 & 340M\\ %335,141,888
			\bottomrule[1.2pt]
		\end{tabular}
		\caption{Statistic of models used. \#L=the number of layers, \#A=the number of attention heads per layer.}
		\label{tab:Models}
	\end{center}
	\vspace{-8mm}
\end{table}

\section{Experiments and Results}

\subsection{Functional Specialization Does Evolve in Multi-head Attention}
% We investigate the degree of functional specialization under dual-task learning and multi-task learning.
\setlength{\tabcolsep}{1.6mm}
\begin{table}[t]
    \small
	\centering
	% \vspace{-2mm}
	\renewcommand\arraystretch{1.0}
	\begin{center}
		\begin{tabular}{cccccc}
			\toprule[1.2pt]
			\textbf{Prune Task}     & \textbf{MNLI}     & \textbf{QQP} 		& \textbf{QNLI}     & \textbf{AG}	    & \textbf{SST-2}    \\
			\midrule[0.8pt]
			MNLI${}^\dagger$        & \underline{58.23} & 71.52 	        & 61.39 	        & 91.99		        & 85.32  	        \\
			QQP${}^\dagger$         & 62.54		        & \underline{69.43} & 60.80 	        & 91.54		        & 85.13 	        \\
			QNLI${}^\dagger$   		& 59.29		        & 70.96 	        & \underline{57.50}	& 91.88		        & 86.35  	        \\
			AG${}^\dagger$          & 65.88		        & 76.28 	        & 69.35 	        & \underline{80.01}	& 85.09 	        \\
			SST-2${}^\dagger$       & 69.50  	        & 77.40		        & 73.96  	        & 86.51		        & \underline{82.45} \\
                \cdashlinelr{1-6}
                Random${}^\dagger$      & 80.68  	        & 85.42		        & 85.05  	        & 93.65		        & 91.23             \\
			\midrule[0.8pt]
			\textbf{Base}           & 83.91  	        & 87.64		        & 90.26  	        & 94.50		        & 92.05    	        \\
                \midrule[0.8pt]
                ${D_i(\alpha)}$           & 7.28  	        & 5.26		        & 11.07  	        & 9.84		        & 3.28               \\
			\bottomrule[1.2pt]
		\end{tabular}
		\caption{Performance(\%) of the pruned and base model on each task using $\text{BERT}_{\text{BASE}}$ with $\alpha=30\%$.
	$\mathcal{T}^\dagger$ denotes top $\alpha$ important heads for this task are pruned. The lowest value is \underline{underlined}.}
		\label{tab:5tasksPrune}
	\end{center}
	\vspace{-8mm}
\end{table}
\setlength{\tabcolsep}{1.5mm}
\begin{table*}[t]
	\centering
	\small
	\renewcommand\arraystretch{1.1}
	\vspace{-0.4cm}
	\begin{center}
		\begin{tabular}{ccccccccccc}
			\toprule[1.2pt]
			      &          &\multicolumn{3}{c}{\textbf{Performance on Task A}}      & \multicolumn{3}{c}{\textbf{Performance on Task B}}  & \\
			 \cmidrule(r){3-5} \cmidrule(r){6-8} \noalign{\smallskip}
			% \multicolumn{1}{c}{Task A}    & \multicolumn{1}{c}{Task B}   & Base Acc.     & Task A${}^\dagger$     & Task B${}^\dagger$   & Base Acc.     & Task A${}^\dagger$     & Task B${}^\dagger$   & $D_{A}(30\%)$       & $D_{B}(30\%)$  & $\overline{D(30\%)}$  \\
			\multicolumn{1}{c}{\textbf{Task A}}    & \multicolumn{1}{c}{\textbf{Task B}}   & Base Acc.     & Task A${}^\dagger$     & Task B${}^\dagger$   & Base Acc.     & Task A${}^\dagger$     & Task B${}^\dagger$   & $D_{A}(30\%)$       & $D_{B}(30\%)$  & $D(30\%)$  \\
			\midrule[0.8pt]
			AG      & QNLI & 94.13         & 85.94         & 92.44             & 91.13         & 65.04             & 52.95             & 6.905                 & 13.270         & 10.088             \\
			AG-Pair & QNLI & 94.46         & 56.17         & 64.52             & 90.72         & 64.85             & 53.66             & 8.842                 & 12.337         & 10.590             \\
                \cdashlinelr{1-11}
                AG      & SST-2 & 94.29         & 89.51         & 92.32             & 92.47         & 89.76             & 86.01             & 2.982                 & 4.051         & 3.517             \\
                AG-Pair & SST-2 & 94.68         & 67.65         & 71.80             & 92.66         & 89.33             & 85.78             & 4.387                 & 3.837         & 4.112             \\
			\bottomrule[1.2pt]
		\end{tabular}
	\end{center}
	\caption{Comparison between different input paradigm combinations. The input of AG-Pair is a pair of sentences from AG, and the label is whether they belong to the same topic. }
	\label{tab:2tasksTrainingParadigm}
 \vspace{-5mm}
\end{table*}
\paragraph{Dual-task Learning}
Based on the pairwise combination of five datasets, there are ten groups of dual-task learning tasks. 
We observe that the dissociation scores of models without frozen in dual-task learning are all positive, i.e., double dissociation phenomenon appears in all task-pairs (details are shown in Appendix \ref{appendix:2tasksWithOtherModels}). 
As illustrated in Figure \ref{fig:average_d_score}, $\text{BERT}_{\text{BASE}}$ shows a distinct functional specialization phenomenon ($D(\alpha)>10\%$) in four dual-task learning tasks. 
Moreover, distinct functional specialization phenomena are also found in the other two sizes of BERT and GPT models.  
The other two base-size models, $\text{RoBERTa}_{\text{BASE}}$ and $\text{DeBERTV3}_{\text{BASE}}$, even show a higher degree of functional specialization, in which average dissociation scores among ten dual-task learning tasks are 13.44\% and 10.88\% respectively.
% (8.80\% in $\text{BERT}_{\text{BASE}}$). 

% As reported in Table \ref{tab:2tasksBertBase}, the dissociation scores of $\text{BERT}_{\text{BASE}}$ in dual-task learning are all positive, i.e., double dissociation phenomenon appears in all task-pairs. 
% The highest one is greater than 30\% and four task pairs show a distinct functional specialization ($\overline{D(\theta)}>10\%$). 
% Moreover, distinct functional specialization phenomena are also found in the other four models investigated in this paper, including different sizes and pre-training methods of multi-head attention-based models (results are shown in Appendix \ref{appendix:2tasksWithOtherModels}).

To eliminate the accidental functional specialization phenomenon, we train another dual-task model using a frozen $\text{BERT}_{\text{BASE}}$ encoder for comparison. 
As shown in the fifth row of Figure \ref{fig:average_d_score}, most of the dissociation scores are relatively small, and only one dual-task pair, ``MNLI and AG'', shows a mild functional specialization phenomenon ($D(\alpha)>5\%$). 
The average dissociation score of these ten task pairs spontaneously increases by 6.32\% if we fine-tune the shared encoder. 
%In other words, the multi-head attention spontaneously segregated itself into distinct subsystems despite the lack of any task-specific inductive bias that might have encouraged this outcome.

\paragraph{Multi-task Learning}
We further conduct multi-task learning experiments using all five tasks in dual-task learning. 
In addition to all positive dissociation scores, we find that the performance of one task decreases more when pruning the top 30\% important attention heads of this task compared with other tasks (Table \ref{tab:5tasksPrune}).
It demonstrates that the functional specialization phenomenon has evolved after multi-task learning, 
i.e., there is a unique group of heads more important to one specific task.
Otherwise, the influence on all tasks would be similar when pruning the most important attention heads for different tasks.

The absolute performances on the first three tasks (MNLI, QQP, and QNLI) suffer a drastic drop after pruning only 30\% attention heads. 
For example, the lowest drop is 14.41\% on MNLI when the top 30\% important heads for SST-2 are pruned, while the highest one is only 14.49\% among the AG and SST-2 when pruning the same amount of attention heads. 
It indicates that tasks taking two sequences as input, e.g., natural language inference and question answering, depend on attention mechanism more than one sequence input task, which is in line with the finding of \citet{vashishth2019attention}.
See Appendix \ref{appendix:5task_results} for more details and analyses. 

\subsection{Task Similarity Affects Functional Specialization}
% \subsection{Exploring Factors Affecting Functional Specialization}
\label{sec:explore_factor}
After observing the functional specialization phenomenon in the multi-head module, it is interesting to study how this phenomenon is affected. 
In this section, we empirically explore two factors: task similarity and input paradigm.
% In this section, we empirically explore three factors: task similarity, input paradigm, and the ratio of training samples.

\begin{table*}[tp]

\renewcommand\arraystretch{0.95}

\centering
\small

\setlength{\tabcolsep}{1.2mm}
 \begin{tabu}{lcccccccccccc}
 
 \toprule[1.2pt]
\multicolumn{1}{c}{\textbf{Model}} & \textbf{Type} & \textbf{\#Params} & \begin{tabular}{@{}c@{}}\textbf{\small CoLA} \\ \scriptsize Mcc\end{tabular} & \begin{tabular}{@{}c@{}}\textbf{\small MNLI-(m/mm)} \\ \scriptsize Acc\end{tabular} & \begin{tabular}{@{}c@{}}\textbf{\small MRPC} \\ \scriptsize F1\end{tabular} & \begin{tabular}{@{}c@{}}\textbf{\small QNLI} \\ \scriptsize Acc\end{tabular} & \begin{tabular}{@{}c@{}}\textbf{\small QQP} \\ \scriptsize F1\end{tabular} & \begin{tabular}{@{}c@{}}\textbf{\small RTE} \\ \scriptsize Acc\end{tabular} & \begin{tabular}{@{}c@{}}\textbf{\small SST-2} \\ \scriptsize Acc\end{tabular} & \begin{tabular}{@{}c@{}}\textbf{\small STS-B} \\ \scriptsize $r^{s}$\end{tabular}& \textbf{\small Avg}\\
   \midrule[0.8pt]
   
$\text{TinyBERT}^{\ddagger}$  & ST & $9.0\times$                & $46.3$          & $83.0/82.4$           & $85.1$             &  $90.0$              & $70.7$          &  $65.6$          & $92.9$           & $84.6$    & $77.8$ \\ 
 \specialrule{0em}{0pt}{0pt}
 
  \cdashlinelr{1-12}
  
$\text{TinyBERT}$  & MTL & $\underline{1.0}\times$              & $35.2$          & $\textbf{82.6}/\textbf{81.9}$           & $83.4$             &  $\textbf{90.5} $              & $70.2$          &  $74.0$          & $92.5$           & $83.5$    & $77.1$ \\ 
 \specialrule{0em}{0pt}{0pt}

 \ \ \ \ \ \ \ \ \ \ \ +IAT  & MTL & $\underline{1.0}\times$    & $\textbf{39.3}$ & $82.5/\textbf{81.9}$           & $\textbf{85.4}$    &  $90.3$     & $\textbf{70.5}$ &  $\textbf{74.1}$ & $\textbf{92.7}$  & $\textbf{84.2}$           & $\textbf{77.9}$ \\ 
 \specialrule{0em}{0pt}{0pt}
 
    \midrule[0.8pt]

$\text{BERT}_{\text{BASE}}^{1}$ & ST & $9.0\times$    & $52.1$   & $84.6/83.4$ & $88.9$          &  $90.5$         & $71.2$          &  $66.4$    & $93.5$            & $85.8$          & $79.6$ \\ 

$\text{PALs}^{2}$               & MTL & $1.13\times$  & $51.2$   & $84.3/83.5$ & $88.7$          &  $90.0$         & $71.5$          &  $76.0$    & $92.6$            & $85.8$          & $80.4$ \\ 

$\text{CA-MTL}_{\text{BASE}}^{3}$ & MTL & $1.12\times$& $53.1$   & $85.9/85.8$ & $88.6$          &  $90.5$         & $69.2$          &  $76.4$    & $93.2$            & $85.3$          & $80.9$ \\ 

 $\text{Ticket-Share}_{\text{BASE}}^{\ddagger}$ & MTL & $\underline{1.0}\times$& $50.3$       & $83.7/83.0$      & $88.0$          &  $90.5$     & $70.5$          &  $76.6$          & $93.7$            & $84.8$          & $80.1$ \\ 

 \specialrule{0em}{0pt}{0pt}
 
 \cdashlinelr{1-12}
 
$\text{BERT}_{\text{BASE}}$ & MTL & $\underline{1.0}\times$& $49.8$ & $\textbf{83.9}/\textbf{83.4}$       & $86.4$          &  $89.9$         & $70.3$          &  $76.0$             & $93.2$            & $85.7$          & $79.8$ \\ 
 \specialrule{0em}{0pt}{0pt}

 \ \ \ \ \ \ \ \ \ \ \ +IAT  & MTL & $\underline{1.0}\times$     & $\textbf{51.6}$     & $\textbf{83.9}/83.1$   & $\textbf{87.6}$ &  $\textbf{90.6}$& $\textbf{71.2}$ &  $\textbf{76.8}$    & $\textbf{94.1}$   & $\textbf{86.2}$ & $\textbf{80.6}$ \\ 
 \specialrule{0em}{0pt}{0pt}
 
    \midrule[0.8pt]
$\text{BERT}_{\text{LARGE}}^{1}$  & ST & $9.0\times$ & $60.5$          & $86.7/85.9$      & $89.3$          &  $92.7$     & $72.1$          &  $70.1$          & $94.9$            & $86.5$            & $82.1$ \\ 
 \specialrule{0em}{0pt}{0pt}

 $\text{Adapters-256}^{4}$        & ST & $1.3\times$ & $59.5$          & $84.9/85.1$      & $89.5$          &  $90.7$     & $71.8$          &  $71.5$          & $94.0$            & $86.9$            & $80.0$ \\ 

 $\text{CA-MTL}_{\text{LARGE}}^{3}$ & MTL & $1.12\times$& $59.5$       & $85.9/85.4$      & $89.3$          &  $92.6$     & $71.4$          &  $79.0$          & $94.7$            & $87.7$          & $82.8$ \\ 
 
 $\text{Ticket-Share}_{\text{LARGE}}^{\ddagger}$ & MTL & $\underline{1.0}\times$& $56.2$       & $86.0/85.6$      & $88.7$          &  $92.7$     & $71.4$          &  $78.8$          & $94.5$            & $85.6$          & $82.2$ \\ 
 
 \specialrule{0em}{0pt}{0pt}
 
 \cdashlinelr{1-12}
 
   $\text{BERT}_{\text{LARGE}}$  & MTL & $\underline{1.0}\times$ & $56.8$          & $\textbf{85.6}/84.9$            & $86.6$                &  $\textbf{92.4}$     & $71.3$          &  $79.0$          & $94.3$            & $86.2$            & $81.9$ \\ 
 \specialrule{0em}{0pt}{0pt}

 \ \ \ \ \ \ \ \ \ \ \ +IAT & MTL & $\underline{1.0}\times$ & $\textbf{60.0}$ & $85.5/\textbf{85.3}$            & $\textbf{88.4}$        &  $92.1$              & $\textbf{71.5}$ &  $\textbf{79.1}$ & $\textbf{94.5}$   & $\textbf{86.8}$   & $\textbf{82.6}$ \\ 
 \specialrule{0em}{0pt}{0pt}
 
    \midrule[0.8pt]
    $\text{RoBERTa}_{\text{BASE}}^{\ddagger}$  & ST & $9.0\times$    & $60.0$          & $87.2/86.7$           & $90.8$             &  $93.1$              & $72.1$          &  $71.9$          & $95.7$           & $88.2$    & $82.9$ \\ 
 \specialrule{0em}{0pt}{0pt}
 
  \cdashlinelr{1-12}
  
   $\text{RoBERTa}_{\text{BASE}}$  & MTL & $\underline{1.0}\times$    & $55.3$          & $\textbf{87.2}/86.7$           & $89.6$             &  $92.3$              & $71.4$          &  $80.0$          & $95.1$           & $\textbf{87.1}$    & $82.7$ \\ 
 \specialrule{0em}{0pt}{0pt}

 \ \ \ \ \ \ \ \ \ \ \ +IAT  & MTL & $\underline{1.0}\times$     & $\textbf{59.9}$ & $86.9/\textbf{86.8}$           & $\textbf{90.9}$    &  $\textbf{92.4}$     & $\textbf{71.8}$ &  $\textbf{80.5}$ & $\textbf{95.4}$  & $\textbf{87.1}$           & $\textbf{83.5}$ \\ 
 \specialrule{0em}{0pt}{0pt}
 
    \midrule[0.8pt]
    
    $\text{DeBERTaV3}_{\text{BASE}}^{\ddagger}$ & ST & $9.0\times$             & $67.1$         & $90.0/89.2$     & $90.6$             &  $94.4$     & $73.9$          &  $81.5$          & $96.2$   & $88.9$              & $85.8$ \\ 
 \specialrule{0em}{0pt}{0pt}

   \cdashlinelr{1-12}

    $\text{DeBERTaV3}_{\text{BASE}}$ & MTL & $\underline{1.0}\times$ & $63.1$         & $\textbf{89.9}/\textbf{89.2}$     & $89.4$             &  $\textbf{93.8}$     & $73.7$          &  $86.7$          & $95.5$   & $89.6$              & $85.7$ \\ 
 \specialrule{0em}{0pt}{0pt}

 \ \ \ \ \ \ \ \ \ \ \ +IAT    & MTL & $\underline{1.0}\times$ & $\textbf{67.2}$  & $89.7/\textbf{89.2}$              & $\textbf{90.9}$    &  $\textbf{93.8}$     & $\textbf{74.0}$ &  $\textbf{86.9}$ & $\textbf{95.8}$   & $\textbf{89.7}$   & $\textbf{86.4}$ \\ 
 \specialrule{0em}{0pt}{0pt}
 
\bottomrule[1.2pt]
\end{tabu}
\caption{\label{tab:glue_test_annealed} GLUE test set results using the GLUE evaluation server. ``ST'' stands for the single task fine-tuned model, whereas ``MTL'' denotes the multi-task learning model. The multi-task learning models we tested are not further fine-tuned on each task, so there is only one model for all tasks ($1.0\times$ in \#Params). Results from: \citet{devlin-etal-2019-bert}${}^1$, \citet{stickland2019pals}${}^2$, \citet{pilault2021conditionally}${}^3$, \citet{houlsby2019Adapter}${}^4$. ${}^{\ddagger}$ indicates our implement result for a fair comparison. The highest performance in the last two conditions of each model is displayed in \textbf{bold}.
}
\vspace{-5mm}
\end{table*}

\paragraph{Task Similarity}
%For a fair comparison, experiments are conducted under a multi-task learning setting due to different amounts of training samples in dual-task learning.
The task similarity metric \textit{Cognitive-Neural Mapping} (CNM), which is found less sensitive to underlying models \citep{luo-etal-2022-cogtaskonomy}, is utilized to quantify the similarity of task-pair in this section.
 
\begin{figure}[t]
\centering 
\setlength{\textfloatsep}{2pt}
\setlength{\intextsep}{2pt}
\setlength{\abovecaptionskip}{2pt}
\includegraphics[width=0.48\textwidth]{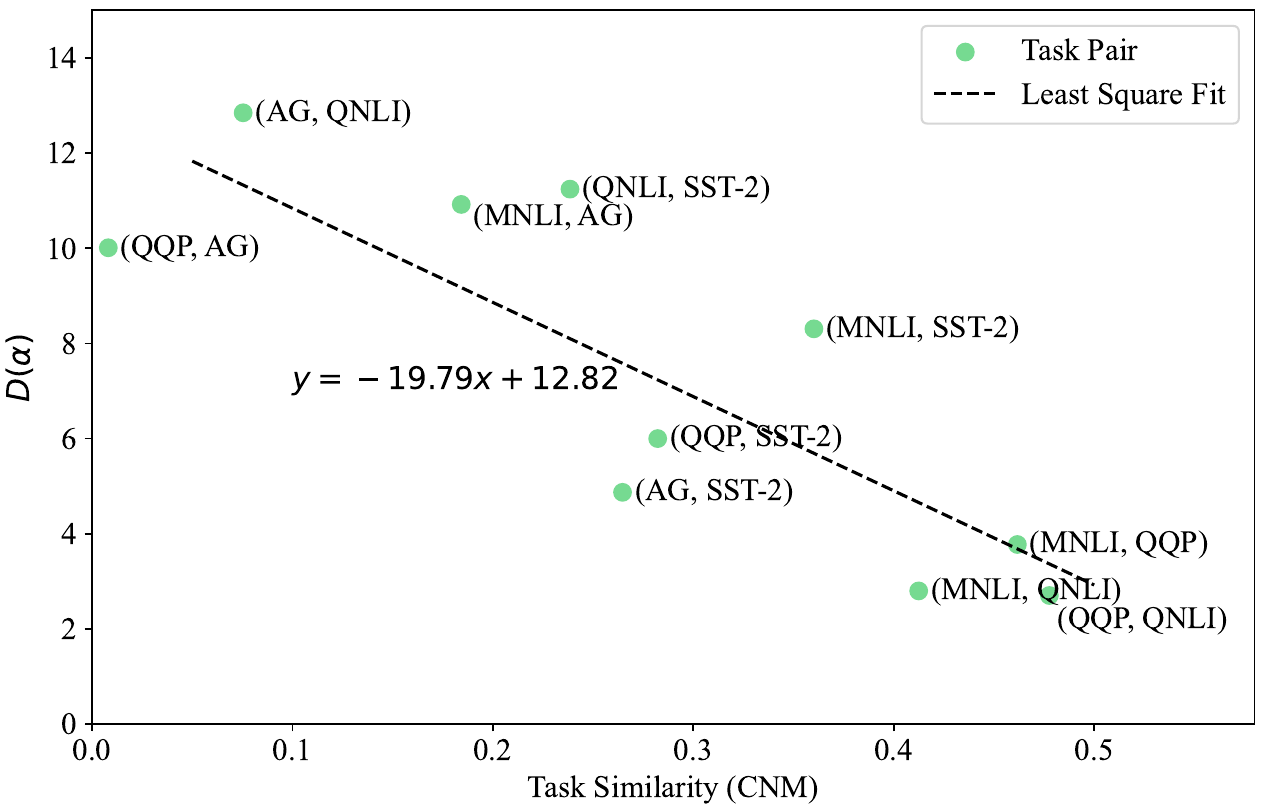}
\caption{The average dissociation score and similarity of each task-pair in multi-task learning. 
}\label{fig:5tasks_average_d}
\vspace{-5mm}
\end{figure}

% The least square method is used to fit the data and showed by the dotted line in Figure \ref{fig:5tasks_average_d}.
As shown in Figure \ref{fig:5tasks_average_d}, we observe that there is a significant negative correlation between the average dissociation score of task-pair and the similarity between tasks. 
In other words, the more similar the tasks are, the lower the average dissociation score is, which suggests the weaker the functional head specialization phenomenon is.
The other three task similarity metrics used and fitting results refer to Appendix \ref{appendix:similarity_results}, where this negative relationship is also found.

% \paragraph{Number of Classes}
% DBPedia and QNLI (Language Model);

% CIFAR-X and AG (Vison-Language Model).

\paragraph{Input Paradigm}
There are two different input paradigms, sentence pair (MNLI, QNLI, and QQP) and single sentence (AG and SST-2), among these five tasks.
We notice the average dissociation score of two tasks in different input paradigms is higher than the same input paradigm ones in Figure \ref{fig:average_d_score} ($\text{BERT}_{\text{BASE}}$: $12.654\%>3.016\%$).
Thus, experiments are conducted to investigate the effect of input paradigm on the degree of functional specialization in multi-head attention.
Specifically, we construct a dataset named ``AG-Pair'' using the sentences of AG dataset, which aims to identify whether a pair of input sentences belong to the same topic. 
The number of samples in AG-Pair is the same as AG, which is 120k, and each sample in AG occurs twice in the AG-Pair dataset.
The generation method and statistics of AG-Pair are reported in Appendix \ref{appendix:AGPair}.

As shown in Table \ref{tab:2tasksTrainingParadigm}, there is no significant dissociation score difference between ``AG + QNLI'' and ``AG-Pair + QNLI'' dual-task learning tasks, which also holds for ``AG + SST-2'' and ``AG-Pair + SST-2''.
We note that the absolute performances on the AG-Pair dataset suffer a drastic drop after pruning only 30\% attention heads, which is similar to the other tasks taking a pair of sentences as input (Table \ref{tab:5tasksPrune}).

According to the experimental results presented above, we observe that task similarity plays a more important role than the input paradigm in the functional specialization of the multi-head attention module. 

\subsection{Improving Multi-Task Models by Training Important Attention Heads}
\label{sec:improve_glue}

Once the importance of attention heads for each task is figured out, we should be able to consolidate their roles by only finetuning them. 
Thus, Important Attention-head Training (IAT) (Section \ref{sec:IAT}) is applied to the multi-task learning models on 9 GLUE datasets and compared against vanilla multi-task learning. 
We observe that the degree of functional specialization in the multi-head attention module is improved by training the top important attention heads during the last part of multi-task learning (details refer to Appendix \ref{appendix:other_glue}).

\begin{table*}[tp]

\renewcommand\arraystretch{1.3}

\centering
\small

\setlength{\tabcolsep}{1.2mm}
 \begin{tabu}{lccccccccc}
 
 \toprule[1.2pt]
   & \multicolumn{4}{c}{\textbf{\# Samples of SciTail}} & \multicolumn{4}{c}{\textbf{\# Samples of IMDB}} \\
  \cmidrule(r){2-5}  \cmidrule(r){6-9} \noalign{\smallskip}
\multicolumn{1}{c}{\textbf{Model}}& \textbf{4}&\textbf{16}& \textbf{32}&\textbf{100}& \textbf{4}&\textbf{16} & \textbf{32}&\textbf{100}\\

   \midrule[0.8pt]
 
 % $\text{MT-DNN}_{\text{SMART}}^{2}$      & $82.3$    & $88.6$           & $91.3$           & $\textbf{96.1}$ &  $82.7$           & $86.0$          &  $\textbf{88.7}$          & $\textbf{91.6}$ \\ 
 
 % $\text{CA-MTL}^{3}$                     & $83.2$    & $88.7$           & $91.4$           & $95.6$          &  $82.8$           & $\textbf{86.2}$ &  $88.0$          & $91.5$ \\

  $\text{MT-DNN}$                  & $71.83_{\pm 6.5}$    & $81.24_{\pm 3.8}$         & $82.59_{\pm 2.3}$           & $85.90_{\pm 2.0}$      & $77.65_{\pm 4.7}$    & $80.76_{\pm 3.1}$      & $82.98_{\pm 0.9}$           & $83.65_{\pm 0.6}$ \\

 $\text{Ticket-Share}_{\text{BASE}}$       & $73.17_{\pm 6.1}$    & $82.07_{\pm 4.0}$         & $83.05_{\pm 2.4}$           & $86.22_{\pm 1.5}$      & $78.43_{\pm 4.0}$    & $81.57_{\pm 1.6}$      & $83.07_{\pm 0.6}$           & $83.84_{\pm 0.5}$ \\
  
  %\midrule[0.8pt]
    \cdashlinelr{1-9}
    
  $\text{BERT}_{\text{BASE}}$            & $69.44_{\pm 8.9}$    & $79.41_{\pm 4.7}$         & $81.52_{\pm 3.0}$           & $85.65_{\pm 1.6}$      & $72.21_{\pm 6.2}$    & $78.67_{\pm 3.5}$      & $82.10_{\pm 1.0}$           & $83.39_{\pm 0.5}$ \\
 \specialrule{0em}{0pt}{0pt}

 \ \ \ \ \ \ \ \ +IAT                    & $\textbf{75.66}_{\pm 4.0}$    & $\textbf{82.11}_{\pm 3.2}$& $\textbf{83.82}^*_{\pm 1.9}$& $\textbf{86.60}^*_{\pm 1.3}$ &  $\textbf{80.50}^*_{\pm 2.6}$& $\textbf{82.03}^*_{\pm 1.9}$ &  $\textbf{83.33}^*_{\pm 0.5}$ & $\textbf{84.08}^*_{\pm 0.3}$ \\  
 \specialrule{0em}{0pt}{0pt}
 
\bottomrule[1.2pt]
\end{tabu}
\caption{\label{tab:transfer_dev} Few-shot transfer learning results on development sets across 30 seeds (${}^{*}$ indicates statistically significant improvements of 5\% level). All models use $\text{BERT}_{\text{BASE}}$ as encoder and are initialized from their multi-task learning models on GLUE. 
}
\vspace{-5mm}

\end{table*}

Table \ref{tab:glue_test_annealed} reports on a comparison result of single task fine-tuning models, multi-task learning models as well as the models using adapters on GLUE test set.\footnote{For a fair comparison, we treat MNLI-m and MNLI-mm as two tasks, which is the same as \citet{houlsby2019Adapter} and \citet{pilault2021conditionally}.}
GPT and GPT-2 are not incorporated due to their inferior performance on GLUE. 
With important attention-head training, the average performances of five multi-task learning models are increased by 0.76\% on average over the vanilla multi-task learning baseline. 
These transformer family models for multi-task learning even surpass their single task fine-tuning counterparts, which consist of 9 task-specific models. 

In most cases, multi-task learning models with IAT receive a performance gain on the four small datasets (CoLA, MRPC, RTE, and STS-B), among which the improvement on CoLA is the most significant (+3.6\% on average).
It comes from the alleviation of negative transfer in CoLA under multi-task learning. 
For example, compared with fine-tuning on CoLA (60.5\%), the performance of $\text{BERT}_{\text{LARGE}}$ drops to 56.8\% under multi-task learning, while it increases to 60.0\% after using IAT. 
The performances of multi-task learning models on two large datasets, QQP and SST-2, are also improved by our method. 
More results, including different sampling methods and performances on GLUE development sets, are shown in Appendix \ref{appendix:other_glue}. 

% In summary, the above results indicate that the degree of functional specialization and generalization ability of multi-task learning models are both improved by IAT.
% The performance improvement in multi-task learning models may arise from this more specialized module.

\paragraph{Few-shot Transfer Learning}
%Transfer knowledge across tasks is an inherent ability of human.
Furthermore, we investigate whether a multi-task learning model with a more specialized multi-head attention module will be better at transfer learning. 
% The model for a new task is initialized from the ones trained on GLUE with multi-task learning except for the last classifier layer.
Table \ref{tab:transfer_dev} presents the few-shot transfer learning results using different amounts of training samples from SciTail (natural language inference task) and IMDB (sentiment analysis task). 
We find that the model initialized from a multi-task learning model using IAT achieves a higher accuracy on the new task, especially when fewer samples are provided. 
IAT degrades to the multi-task learning method proposed by \citet{liang-etal-2021-super} when $\delta = 1$, and often obtains a worse performance in multi-task learning and transfer learning (Ticket-Share in Table \ref{tab:glue_test_annealed} and Table \ref{tab:transfer_dev}). 
It may come from the weak functional specialization phenomenon in the original pre-trained models (e.g., the frozen $\text{BERT}_{\text{BASE}}$ encoder in Figure \ref{fig:average_d_score}), which makes it harder to correctly determine the most important attention heads for each task at the beginning of multi-task training.

\paragraph{Ablation Study}
To take a deep look into the improvements contributed by important attention-head training, we conduct an ablation study on GLUE dev set using $\text{BERT}_{\text{BASE}}$ (Table \ref{tab:ablation_IAT}). 
After pruning the least important 30\% heads, there is a performance gain on three tasks (MRPC, SST-2, and STS-B), which is in line with the previous finding that Transformer can be improved by pruning some redundant attention heads \citep{Michel2019Are16HeadsBetter}.

It is interesting to find that multi-task models can benefit from training random 30\% attention heads for each task, which may arise from the mitigation of gradient interference by subdividing the parameters shared. 
Compared with training random 30\% attention heads, training the most important part of attention heads can further improve the average performance and benefit more tasks, which is confirmed in the ablation study of few-shot transfer learning (Table \ref{tab:ablation_transfer_IAT}). 

\setlength{\tabcolsep}{0.8mm}
\begin{table}[t]
	\centering
	\footnotesize
	%\vspace{-0.4cm}
	\renewcommand\arraystretch{1.2}
	\begin{center}
		\begin{tabular}{lcc}
			\toprule[1.2pt]  
			\multicolumn{1}{c}{\textbf{Model}}                & \textbf{Avg}                 &  {\scriptsize \textbf{\# Tasks Improved}} \\
			\midrule[0.8pt]
			\multicolumn{1}{l}{$\text{BERT}_{\text{BASE}}$}   & $82.64_{\pm 0.09}$           &   -  \\
                \midrule[0.8pt]
                {\scriptsize w/ Prune the least important 30\% heads}         & $82.23_{\pm 0.46}$           &   3  \\
                {\scriptsize w/ Train random 30\% heads}         & $82.71_{\pm 0.10}$           &   6  \\
			{\scriptsize w/ Train the most important 30\% heads}              & $\textbf{83.41}_{\pm 0.20}$  &   8  \\
			\bottomrule[1.2pt]
		\end{tabular}
		\caption{Ablation study of different multi-task methods on GLUE dev set with $\delta = 10\%$. }
		\label{tab:ablation_IAT}
	\end{center}
	\vspace{-3mm}
\end{table}
\setlength{\tabcolsep}{0.8mm}
\begin{table}[t]
	\centering
	\footnotesize
	%\vspace{-0.4cm}
	\renewcommand\arraystretch{1.2}
	\begin{center}
		\begin{tabular}{lcc}
			\toprule[1.2pt]  
			\multicolumn{1}{c}{\textbf{Model}}                & \textbf{SciTail}                 &  \textbf{IMDB} \\
			\midrule[0.8pt]
			\multicolumn{1}{l}{$\text{BERT}_{\text{BASE}}$}   & $69.44_{\pm 8.94}$           &   $72.21_{\pm 6.17}$  \\
                \midrule[0.8pt]
                {\scriptsize w/ Train random 30\% heads}         & $72.61_{\pm 7.13}$           & $77.81_{\pm 3.76}$   \\
			{\scriptsize w/ Train the most important 30\% heads}              & $\textbf{75.66}_{\pm 4.02}$  &   $\textbf{80.50}_{\pm 2.63}$  \\
			\bottomrule[1.2pt]
		\end{tabular}
		\caption{Ablation study of 4-shot transfer learning using different multi-task learning models on GLUE. }
		\label{tab:ablation_transfer_IAT}
	\end{center}
	\vspace{-5mm}
\end{table}

% experiments: 8 tasks on GLUE, improve average performance and decrease the standard deviation among performance (BERT-base). 
% [Intuitively, when negative task transfer occurs between two tasks, either (1) task interference is bidirectional and scores are both impacted, or (2) interference is unidirectional and only one score is impacted.]

% \begin{figure}[t]
% \centering 
% \setlength{\textfloatsep}{2pt}
% \setlength{\intextsep}{2pt}
% \setlength{\abovecaptionskip}{2pt}
% \includegraphics[width=3in]{imgs/acc_on_scitail.png}
% \caption{Accuracy on Scitail development set using 1\% training samples. 
% }\label{fig:acc_scitail}
% \vspace{-4mm}
% \end{figure}

\section{Conclusions and Future Work}
% do not just repeat the content in abstract. there should be more insightful conclusions and takes away for the audience. maybe some future work. 
% The research on functional specialization in the human brain shed light on its learning and processing mechanism. 
% Therefore, to better understand the Transformer-based models, this paper investigates the brain-like functional specialization phenomenon in multi-head attention. 
In this paper, we conduct extensive dissociation experiments and observe that the brain-like functional specialization phenomenon does evolve in multi-head attention after dual-task or multi-task learning. 
Furthermore, experimental results show that the performance and generalization ability of multi-task models can be improved by the multi-task training method based on functional specialization.  
This work, inspired by neuroscience findings, studies the interpretation and improvement of neural networks, which we hope will promote more efforts on interdisciplinary work combining neuroscience and artificial intelligence. 

% 解释更多的模块，以及更好的利用方法
In the future, we plan to investigate more neural network modules that may arise the functional specialization phenomenon under multi-task learning. 
% Another direction is to interpret the important attention heads for each task under multi-task learning. 
Another direction is to design better methods exploiting this phenomenon to further improve multi-task learning models. 
% 可以删了
% For example, ten probing tasks, requiring surface, syntactic and semantic information, can be utilized to interpret the sentence representations from attention heads. 
% We observe that the top important attention heads for entailment tasks perform better on semantic probing tasks, whereas the ones for topic classification tasks are better on surface tasks (details in Appendix \ref{appendix:5task_probe_interpret}).

\section*{Limitations}
% this first point is not a limitation, it is more like a potential extension. I don't understand the second one.

% 我们现在的任务还仅限于理解任务
Firstly, we conduct extensive experiments on multiple natural language understanding tasks only, and multi-modal tasks could be investigated further. 

In addition, only one approach is utilized to estimate the importance of each attention head, and the most important attention heads are pruned at once. 
Because of this choice, our results can be seen as a lower bound on the estimation of functional specialization in multi-head attention. 
We acknowledge that there might be methods to show higher dissociation scores, such as adopting other attention head importance estimation methods \citep{hao-et-al-2021-interpreting-information,li-etal-2021-differentiable} or iterative pruning. 

We note that the four similarity metrics used in this study are model-dependent, and recognize that results might be different for other Transformer-based models.

Lastly, there are two hyper-parameters introduced in our multi-task training method, which may need extra tuning when adapted to other multi-task learning settings. 

% We only investigate five Transformer-based models in this study, which can be easily applied to other models based on multi-head attention, for example, XLNet \citep{yang2019xlnet}. 
% % We conduct dissociation experiments using only natural language processing tasks, and more real-world tasks, such as computer vision and speech recognition tasks, could be investigated further. 

% Another limitation of this work arises from only an approach applied to estimate the importance of attention head, which can be replaced and compared with other methods like the Self-Attention Attribution Method introduced in \citet{hao-et-al-2021-interpreting-information}. 

% Note the similarity metrics used in this study are model-dependent, which means the negative correlation between task similarity and functional specialization may not be applied to the other models.

% \section*{Ethics Statement}
% Scientific work published at ACL 2023 must comply with the ACL Ethics Policy.\footnote{\url{https://www.aclweb.org/portal/content/acl-code-ethics}} We encourage all authors to include an explicit ethics statement on the broader impact of the work, or other ethical considerations after the conclusion but before the references. The ethics statement will not count toward the page limit (8 pages for long, 4 pages for short papers).

\section*{Acknowledgements}
We would like to thank the anonymous reviewers for their valuable comments. 
The research work was supported by the National Science Foundation of China (No. 62122088) and STI2030-Major Project (No. 2021ZD0204105).

\bibliography{anthology,custom}
\bibliographystyle{acl_natbib}

\appendix
\newpage
\section{Hyperparameters}
\label{appendix:hyper_parameters}
\subsection{Dual-task and Multi-task Learning}
To fine-tune the pre-trained models on dual-task or multi-task learning, we use Adam optimizer \cite{Kingma2015adam}, in which $\beta_1 = 0.9$ and $\beta_2 = 0.999$, and a learning rate of 2e-5. 
We also use a linear warm-up schedule and set the warm-up proportion to 0.1.
The number of epochs is empirically set to 5 for a fair comparison. 
The only exception is the distillation of TinyBERT, which contains \textit{intermediate layer distillation} and \textit{prediction layer distillation}. 
Under the supervision of a fine-tuned $\text{BERT}_{\text{BASE}}$, these distillation methods are performed for 2 and 3 epochs without augmented data, respectively.
Unless otherwise specified, the proportional sampling method is utilized in multi-task learning. 

Similar to the difference in area between cortical regions, the best $\alpha$ for each task may be different in dissociation experiments. 
We acknowledge that higher dissociation scores can be obtained by fine-tuning $\alpha$ in each dual-task learning task. 
For a fair comparison, $\alpha$ is empirically set to 30\% in all dissociation experiments to show the extent of functional specialization in the multi-head attention module. 
All experiments are repeated under three random seeds and average results are reported. 

% \subsection{Probing Tasks}
% \label{appendix:probe_tasks}
% \input{tabs/probingTasks.tex}
% The SentEval toolkit is utilized for surface, syntactic, and semantic probing tasks \footnote{\url{https://github.com/facebookresearch/SentEval}}.
% Table \ref{tab:probingTasks} shows the statistic of ten probing tasks.
% We adopt logistic regression layer (nhid=0), which takes a sentence representation vector as input, to obtain the results of the probing task.
% This classifier is trained using a RMSProp optimizer with 0.1 learning rate, 256 batch size, and tested with 5-fold cross validation. 
% We only tune L2 regularization amongst $\{$10-5, 10-4, 10-3, 10-2, 10-1$\}$ and stop training when the validation accuracy has not improved over the last three epochs (tenacity=3).

\subsection{Transfer Learning}
\label{appendix:statistic_significance}
Since only a small part of training samples are used in transfer learning experiments, we increase the number of training epochs to 20, and conduct a paired bootstrap statistical test under 30 random seeds \cite{dror-etal-2018-hitchhikers}. 

\section{Dual-task Learning Experiments}
\label{appendix:2tasksWithOtherModels}
In this section, we present the results of all multi-head attention based models investigated in dual-task learning tasks.

As reported in Table \ref{tab:2tasks_all}, the dissociation scores of Transformer-based models in dual-task learning are all positive when fine-tuning the pre-trained encoder, i.e., double dissociation phenomenon appears in all task-pairs. 
% The highest one is greater than 30\%
It further demonstrates that the functional specialization phenomenon does appear in the multi-head attention module after training on these dual-task learning tasks.

% Similar to $\text{BERT}_{\text{BASE}}$, distinct functional specialization phenomena do appear in TinyBERT (Table \ref{tab:2tasksBertTiny}) and $\text{BERT}_{\text{LARGE}}$ (Table \ref{tab:2tasksBertLarge}) under dual-task learning settings.

% The other two base-size models, $\text{RoBERTa}_{\text{BASE}}$ (Table \ref{tab:2tasksRoberta}) and $\text{DeBERTV3}_{\text{BASE}}$ (Table \ref{tab:2tasksDebertaV3}), even show a higher degree of functional specialization, in which average dissociation scores are 13.44\% and 10.88\% respectively (8.80\% in $\text{BERT}_{\text{BASE}}$). 

\begin{table*}[tp]

\renewcommand\arraystretch{1.2}

\centering
\scriptsize

\setlength{\tabcolsep}{0.4mm}
 \begin{tabu}{ccc|cc|cc|cc|cc|cc|cc|cc|cc|cc}
 
 \toprule[1.2pt]
    &  \multicolumn{8}{c}{$\text{MNLI}_A$} & \multicolumn{6}{c}{$\text{QQP}_A$} & \multicolumn{4}{c}{$\text{QNLI}_A$} & \multicolumn{2}{c}{$\text{AG}_A$} \\
  \cmidrule(r){2-9} \cmidrule(r){10-15} \cmidrule(r){16-19} \cmidrule(r){20-21}  \noalign{\smallskip}
    &  \multicolumn{2}{c}{$\text{QQP}_B$} & \multicolumn{2}{c}{$\text{QNLI}_B$} & \multicolumn{2}{c}{$\text{AG}_B$}  & \multicolumn{2}{c}{$\text{SST-2}_B$} & \multicolumn{2}{c}{$\text{QNLI}_B$} & \multicolumn{2}{c}{$\text{AG}_B$}  & \multicolumn{2}{c}{$\text{SST-2}_B$}  & \multicolumn{2}{c}{$\text{AG}_B$}  & \multicolumn{2}{c}{$\text{SST-2}_B$}  & \multicolumn{2}{c}{$\text{SST-2}_B$}   \\
  \cmidrule(r){2-3} \cmidrule(r){4-5} \cmidrule(r){6-7} \cmidrule(r){8-9} \cmidrule(r){10-11} \cmidrule(r){12-13} \cmidrule(r){14-15} \cmidrule(r){16-17} \cmidrule(r){18-19} \cmidrule(r){20-21}   \noalign{\smallskip}
  
\multicolumn{1}{c}{\textbf{Model}} & $D_A$ & $D_B$ & $D_A$ & $D_B$ & $D_A$ & $D_B$ & $D_A$ & $D_B$ & $D_A$ & $D_B$ & $D_A$ & $D_B$ & $D_A$ & $D_B$ & $D_A$ & $D_B$ & $D_A$ & $D_B$ & $D_A$ & $D_B$ \\

   \midrule[0.8pt]
   
$\text{GPT}$ & $0.28$ & $3.76$ & $2.10$ & $1.36$ & $16.75$ & $10.47$ & $16.49$ & $1.13$ & $1.36$ & $0.80$ & $18.19$ & $3.60$ & $25.19$ & $1.08$ & $28.74$ & $3.51$ & $31.34$ & $0.75$ & $2.51$ & $4.08$ \\

 $\text{GPT-2}$ & $4.76$ & $3.26$ & $3.09$ & $0.71$ & $15.44$ & $3.29$ & $17.69$ & $5.79$ & $12.64$ & $0.66$ & $19.24$ & $2.39$ & $50.19$ & $4.18$ & $18.62$ & $4.69$ & $16.49$ & $6.81$ & $3.60$ & $6.61$ \\

   \midrule[0.8pt]
   
   $\text{TinyBERT}$ & $2.04$ & $20.10$ & $3.82$ & $1.19$ & $35.66$ & $5.36$ & $31.85$ & $4.71$ & $14.42$ & $0.63$ & $8.70$ & $6.54$ & $24.39$ & $2.35$ & $14.87$ & $6.28$ & $35.59$ & $1.63$ & $1.08$ & $1.59$ \\

 $\text{BERT}_{\text{BASE}}$ & $2.78$ & $2.60$ & $2.34$ & $2.63$ & $11.36$ & $16.26$ & $13.81$ & $0.21$ & $1.92$ & $4.82$ & $28.97$ & $10.39$ & $11.34$ & $3.18$ & $13.27$ & $6.91$ & $32.87$ & $3.28$ & $2.98$ & $4.05$ \\

 \textcolor{myblue}{$\text{BERT}_{\text{BASE}}^{*}$} & $-0.96$ & $0.35$ & $-0.13$ & $1.21$ & $5.59$ & $4.53$ & $3.27$ & $1.20$ & $0.01$ & $0.15$ & $0.82$ & $3.94$ & $-1.07$ & $6.95$ & $6.74$ & $0.90$ & $7.68$ & $2.14$ & $2.48$ & $3.90$ \\ 

 $\text{BERT}_{\text{LARGE}}$ & $2.43$ & $0.83$ & $5.69$ & $3.55$ & $17.31$ & $6.16$ & $19.83$ & $9.81$ & $0.02$ & $5.90$ & $22.40$ & $2.88$ & $21.20$ & $6.52$ & $17.90$ & $9.56$ & $20.44$ & $8.73$ & $7.72$ & $7.23$  \\ 
 
   \midrule[0.8pt]
   
 $\text{RoBERTa}_{\text{BASE}}$ & $8.56$ & $10.86$ & $1.31$ & $4.32$ & $17.43$ & $38.23$ & $5.48$ & $8.40$ & $2.43$ & $0.54$ & $23.27$ & $22.73$ & $22.19$ & $5.59$ & $15.96$ & $47.73$ & $16.09$ & $5.01$ & $6.62$ & $6.16$ \\ 

 $\text{DeBERTaV3}_{\text{BASE}}$ & $6.21$ & $4.64$ & $14.21$ & $0.02$ & $24.11$ & $8.47$ & $27.03$ & $2.65$ & $9.50$ & $0.02$ & $11.65$ & $9.50$ & $20.51$ & $13.33$ & $3.42$ & $16.92$ & $20.36$ & $2.33$ & $4.97$ & $17.78$ \\ 

\bottomrule[1.2pt]

\end{tabu}
\caption{\label{tab:2tasks_all} Results in dual-task learning experiments under $\alpha = 30\%$. \textcolor{myblue}{${}^*$} indicates that the parameters of $\text{BERT}_{\text{BASE}}$ encoder are \textcolor{myblue}{\textbf{frozen}}.
}.
\vspace{-5mm}
\end{table*}

% \input{tabs/dual_task/bert_tiny.tex}

% \input{tabs/dual_task/bert_base.tex}

% \input{tabs/dual_task/bert_baseline.tex}

% \input{tabs/dual_task/bert_large.tex}

% \input{tabs/dual_task/roberta_base.tex}

% \input{tabs/dual_task/debertaV3.tex}

% 可能不太吻合
% \begin{figure}[t]
% \centering 
% \setlength{\textfloatsep}{2pt}
% \setlength{\intextsep}{2pt}
% \setlength{\abovecaptionskip}{2pt}
% \includegraphics[width=0.45\textwidth]{imgs/ratio_MNLI-AG.png}
% \caption{The dissociation scores under different ratios of training samples in dual-task learning with $\theta = 30\%$. 
% }\label{fig:mnli_ag_ratio}
% \vspace{-2mm}
% \end{figure}

% \paragraph{Ratio of Training Samples}

% We investigate the effect of different ratios of training samples used in ``MNLI + AG'' dual-task learning.
% The ratio of training samples between tasks is altered by randomly sampling different amounts of MNLI samples while keeping the number of AG samples unchanged, which is 120k. 
% Figure \ref{fig:mnli_ag_ratio} illustrates the dissociation scores for each task under different ratios of training samples. 
% It has a small impact on the average dissociation score, which is less than 4\%, and the distinct functional specialization phenomenon still appears under all ratios tested. 

\section{Multi-task Learning Experiments on $\text{BERT}_{\text{BASE}}$}
\label{appendix:5task_results}
We report more results of multi-task learning experiments conducted in Section \ref{sec:explore_factor}.
The pair-wise dissociation scores are reported in Table \ref{tab:5tasksDscores}. 
\setlength{\tabcolsep}{1.7mm}
\begin{table}[ht]
	\centering
	\small
	\renewcommand\arraystretch{1.3}
	% \vspace{-0.4cm}
	\begin{center}
		\begin{tabular}{cccccc}
			\toprule[1.2pt]
			\textbf{Task}                   & $\text{MNLI}_A$   & $\text{QQP}_A$    & $\text{QNLI}_A$ 	& $\text{AG}_A$     & $\text{SST-2}_A$ 	\\
			\midrule[0.8pt]
			$\text{MNLI}_B$                 & -  		        & 2.385  	        & 4.321 	        & \textbf{12.671}	& 3.115  	        \\
			$\text{QQP}_B$                  & 5.161	            & - 		        & \underline{3.657} & 12.198	        & 2.907 	        \\
			$\text{QNLI}_B$                 & \underline{1.275}	& \underline{1.746} & - 		        & 12.555	        & \textbf{4.236}  	\\
			$\text{AG}_B$                   & 9.169	            & 7.819 	        & 13.135 	        & - 		        & \underline{2.865} \\
			$\text{SST-2}_B$                & \textbf{13.496}   & \textbf{9.091} 	& \textbf{18.246}   & \underline{6.874}	& -        	        \\
			\bottomrule[1.2pt]
		\end{tabular}
		\caption{$D_A(\alpha)$ between task-pairs, which is calculated on the pruning results of multi-task learning with $\alpha=30\%$. The highest dissociation score in each task A is displayed in \textbf{bold}, and the lowest one is \underline{underlined}.}
		\label{tab:5tasksDscores}
	\end{center}
	\vspace{-5mm}
\end{table}

\begin{figure}[ht]
\centering 
\setlength{\textfloatsep}{2pt}
\setlength{\intextsep}{2pt}
\setlength{\abovecaptionskip}{2pt}
\includegraphics[width=0.45\textwidth]{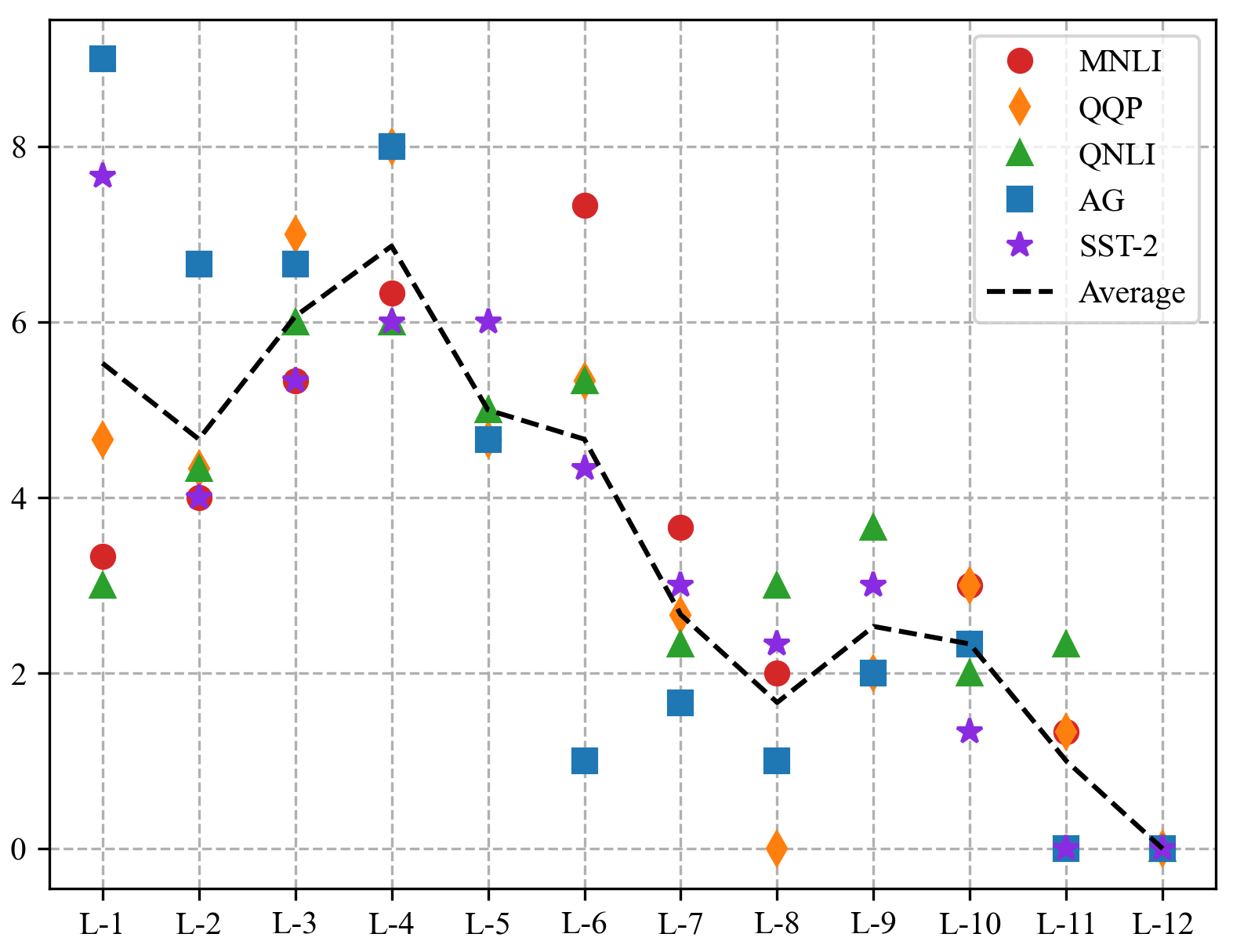}
\caption{The number of important heads pruned among the layers of $\text{BERT}_{\text{BASE}}$ after multi-task learning. The average number of heads pruned in one layer is 3.6 ($\alpha=30\%$).}\label{fig:prune_distribution}
\vspace{-2mm}
\end{figure}

\paragraph{Distribution of Heads Pruned} To gain more insights about the functional specialization in multi-task learning, we statistic the distribution of heads pruned for each task across layers in multi-task learning (Figure \ref{fig:prune_distribution}). 
The average number of attention heads pruned shows a trend of increasing first and then decreasing, which changes at the 4th layer.
The two layers with the greatest difference among tasks are the first layer ($\sigma=2.39$) and the sixth layer ($\sigma=2.08$) of $\text{BERT}_{\text{BASE}}$ after fine-tuning 5 epochs on these five tasks.

\setlength{\tabcolsep}{1.8mm}
\begin{table}[ht]
    \small
	\centering
	% \vspace{-4mm}
	\renewcommand\arraystretch{1.3}
	\begin{center}
		\begin{tabular}{cccccc}
			\toprule[1.2pt]
			\textbf{Task}       & $\text{MNLI}_A$   & $\text{QQP}_A$ 	& $\text{QNLI}_A$ 	& $\text{AG}_A$  	& $\text{SST-2}_A$    \\
			\midrule[0.8pt]
			$\text{MNLI}_B$     & -                 & \textbf{81.40} 	& \textbf{83.70} 	& \underline{58.14}	& \textbf{68.99}    \\
			$\text{QQP}_B$      & 81.40		        & -                 & 78.29 	        & 63.57	            & 62.79             \\
			$\text{QNLI}_B$   	& \textbf{83.70}	& 78.29 	        & -	                & 62.02		        & \underline{62.02} \\
			$\text{AG}_B$       & \underline{58.14}	& 63.57 	        & \underline{62.02} & -	                & 64.34 	        \\
			$\text{SST-2}_B$    & 68.99  	        & \underline{62.79}	& \underline{62.02} & \textbf{64.34}	& -                 \\
			\bottomrule[1.2pt]
		\end{tabular}
	\end{center}
	\caption{The overlapping percentage of important heads pruned in multi-task learning under $\alpha=30\%$. The highest overlapping in each task A is displayed in \textbf{bold}, and the lowest one is \underline{underlined}.}
	\label{tab:head_overlapping}
	\vspace{-1mm}
\end{table}
\paragraph{Overlapping of Heads Pruned}
Table \ref{tab:head_overlapping} reports the overlapping of attention heads pruned between tasks.
It seems that the proportion of overlapping heads pruned does not completely correspond to the dissociation score of each task (Table \ref{tab:5tasksDscores}). 
For example, as for the MNLI and AG tasks, the task with the highest overlapping of heads pruned is the same as the one with the lowest dissociation score. 
However, the highest overlapping of heads pruned for SST-2 comes to the second-highest dissociation score when combined with MNLI.

% \input{tabs/InterpretingImportantHeads.tex}
% \paragraph{Interpretating by Probing Tasks}
% \label{appendix:5task_probe_interpret}
% We attempt to interpret the top important attention heads for each task by ten probing tasks that cover the surface, syntactic and semantic information \citep{conneau-kiela-2018-senteval}. 
% The statistics and hyperparameters of probing tasks refer to Appendix \ref{appendix:probe_tasks}.
% Given a sentence, the hidden state of [CLS] in the output of the attention head is used as sentence representation and utilized for classification in each probing task.

% Table \ref{tab:probing_performance} shows the performance of the top-30\% important attention heads for each task. 
% The top-30\% important attention heads for AG, a topic classification task, perform much better on two surface probing tasks. 
% However, the top-30\% important attention heads for entailment tasks, e.g., MNLI and QNLI, are better on the last four semantic probing tasks. 
% It may come from the different information needed to solve the topic classification tasks and entailment tasks. 

\section{Task Similarity Metrics and Fitting Results}
\begin{figure}[ht]
\centering 
\setlength{\textfloatsep}{2pt}
\setlength{\intextsep}{2pt}
\setlength{\abovecaptionskip}{2pt}
\includegraphics[width=0.48\textwidth]{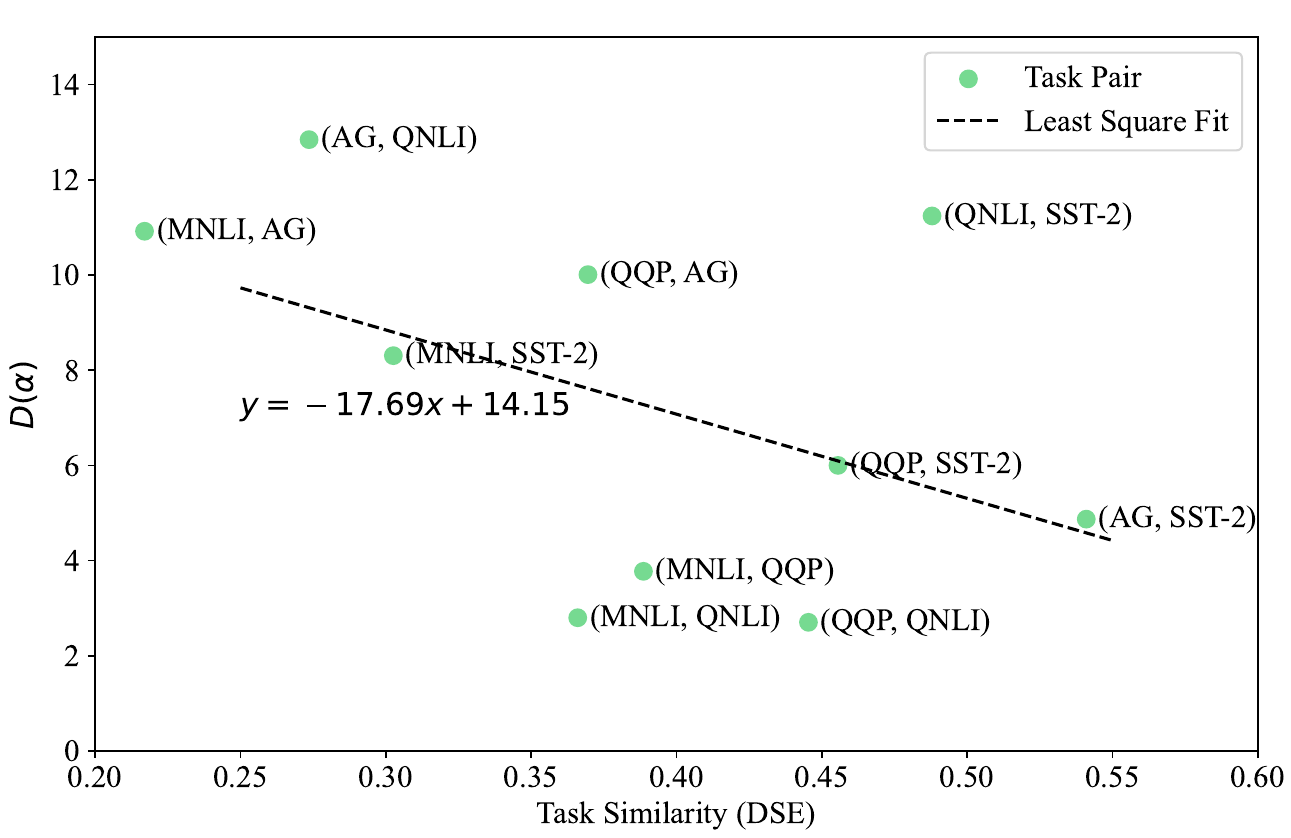}
\caption{The average dissociation score and DSE similarity of each task pair in multi-task learning. 
}\label{fig:5tasks_average_dse}
\vspace{-3mm}
\end{figure}

To verify the robustness of our finding in Section \ref{sec:explore_factor}, the following four metrics are adopted to determine the similarity of each task pair:

\label{appendix:similarity_results}
\paragraph{Direct Similarity Estimation (DSE)} This method approximates the similarity of task pairs by the average similarity of sentence representations from models fine-tuning on the corresponding task. 
Therefore, we randomly select 1000 sentences from the Wikipedia corpus and adopt cosine similarity to quantify the similarity of sentence representations. 
Results with DSE metric are shown in Figure \ref{fig:5tasks_average_dse}. 

\begin{figure}[ht]
\centering 
\setlength{\textfloatsep}{2pt}
\setlength{\intextsep}{2pt}
\setlength{\abovecaptionskip}{2pt}
\includegraphics[width=0.48\textwidth]{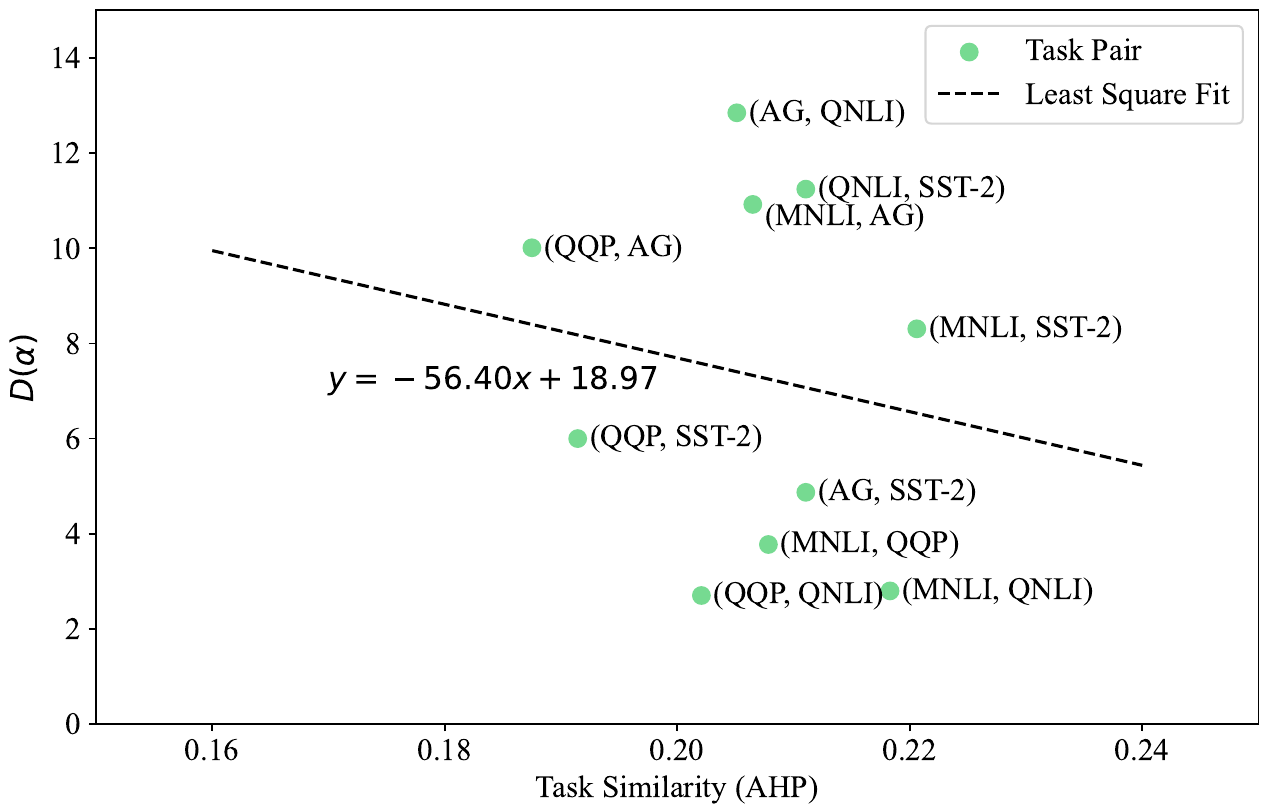}
\caption{The average dissociation score and AHP similarity of each task pair in multi-task learning. 
}\label{fig:5tasks_average_ahp}
\vspace{-3mm}
\end{figure}

\paragraph{Analytic Hierarchy Process (AHP)} On the other hand, the similarity of task pairs can be approximated from the pair-wise transfer learning results \citep{zamir_2018_taskonomy}.
Given a target task, models transferred from different source tasks are compared on a hold-out dataset to determine the transferability of the target task, which is further used to approximate the similarity between tasks. 
Results using AHP are illustrated in Figure \ref{fig:5tasks_average_ahp}. 

\begin{figure}[ht]
\centering 
\setlength{\textfloatsep}{2pt}
\setlength{\intextsep}{2pt}
\setlength{\abovecaptionskip}{2pt}
\includegraphics[width=0.48\textwidth]{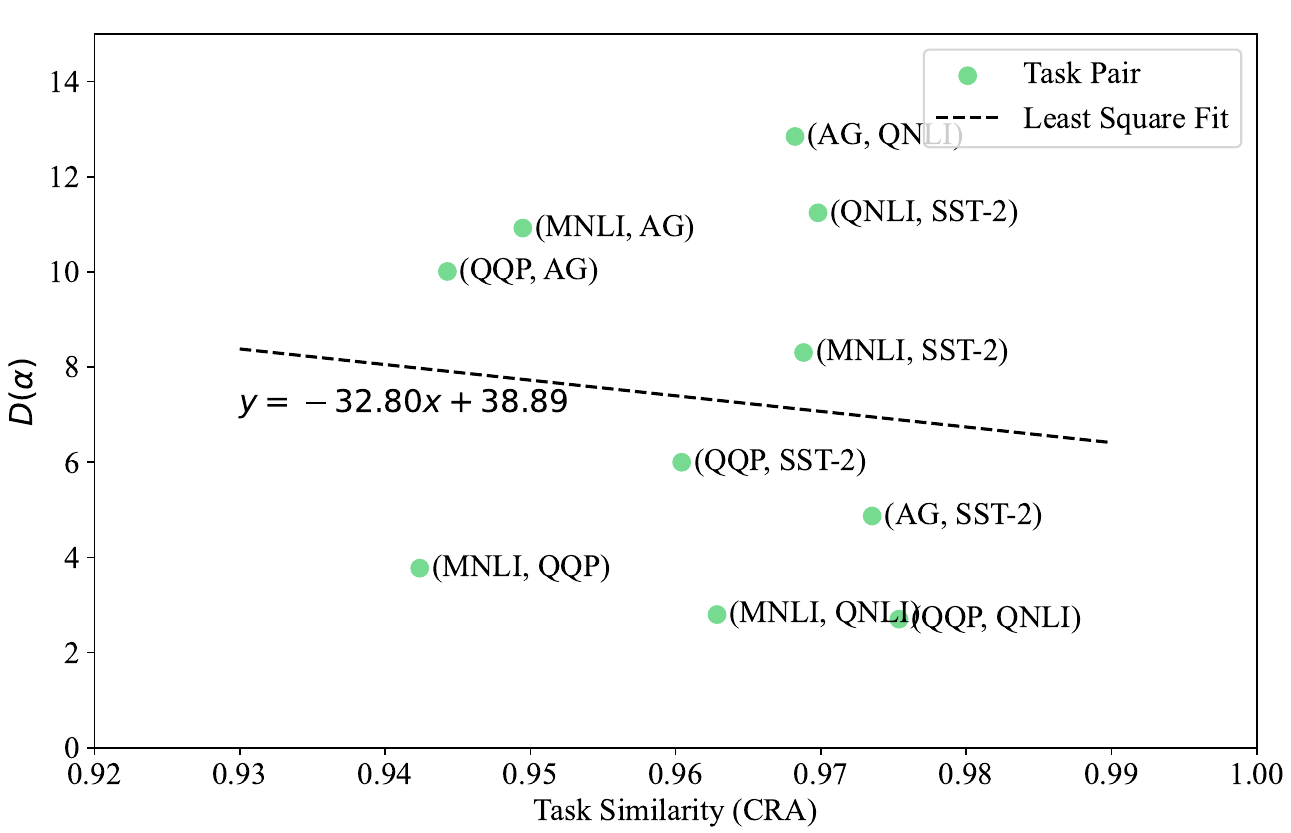}
\caption{The average dissociation score and CRA similarity of each task pair in multi-task learning. 
}\label{fig:5tasks_average_cra}
\vspace{-3mm}
\end{figure}

\paragraph{Cognitive Representation Analytics (CRA)} Inspired by Representational Similarity Analysis (RSA) in cognitive neuroscience \citep{nikolaus-et-al-2008-rsa}, CRA first calculates the Representation Dissimilarity Matrix (RDM) by the dissimilarity of sentence representations, then approximates the similarity between tasks by the similarity between the corresponding RDMs \cite{luo-etal-2022-cogtaskonomy}. 
Figure \ref{fig:5tasks_average_cra} presents the results with CRA.

\paragraph{Cognitive-Neural Mapping (CNM)}
CNM calculates the task similarity by mapping sentence representations of fine-tuned models to fMRI data \citep{luo-etal-2022-cogtaskonomy}, which is recorded when 5 participants were intently reading presented 384 passages \citep{pereira2018toward}. 
Different from randomly selecting 25k fMRI voxels, the most informative 5k fMRI voxels for each participant are used to predict the similarity among tasks. 
Results with CNM have been shown in Figure \ref{fig:5tasks_average_d}. 

To sum up, we observe that there is a negative correlation between the average dissociation score and the task similarity, no matter which task similarity metric is adopted.

\section{AG-Pair dataset}
\label{appendix:AGPair}
The AG-Pair dataset is built from the original dataset AG's News that contains 120k training samples from four topics.
Given a pair of news as input, the model has to predict whether they are belonging to the same topic (Same) or not (Different). 

To generate this dataset, samples in AG are iterated in random order and have an equal chance to combine a sample in the same topic or the other three topics. 
Thus the numbers of training samples in two classes are both 60k. 
Moreover, each news in AG's News occurs exactly twice in the AG-Pair dataset to keep the same word frequency.

\newpage

\section{Other Experimental Results on GLUE}
\label{appendix:other_glue}
\begin{figure}[ht]
\centering 
\setlength{\textfloatsep}{2pt}
\setlength{\intextsep}{2pt}
\setlength{\abovecaptionskip}{2pt}
\includegraphics[width=3in]{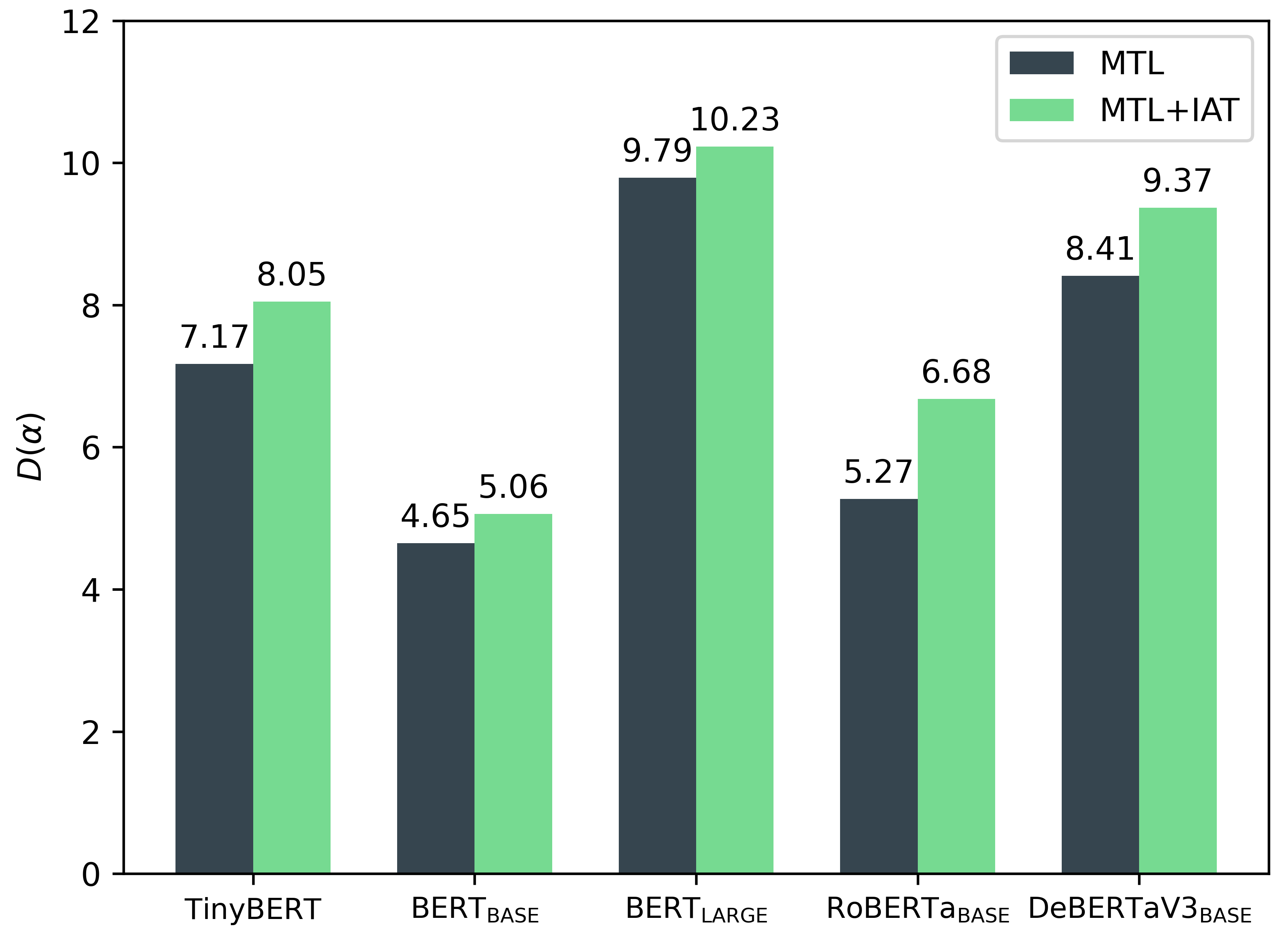}
\caption{Average dissociation score of five multi-task learning models on GLUE dev set with $\alpha=30\%$.
}\label{fig:avg_D_IAT}
\vspace{-5mm}
\end{figure}

In this section, we report more results and analyses of multi-task learning models on GLUE. 
Figure \ref{fig:avg_D_IAT} illustrates that the average dissociation scores of five Transformer-based models are all improved by IAT as we expected. 

\begin{figure}[ht]
\centering 
\setlength{\textfloatsep}{2pt}
\setlength{\intextsep}{2pt}
\setlength{\abovecaptionskip}{2pt}
\includegraphics[width=0.45\textwidth]{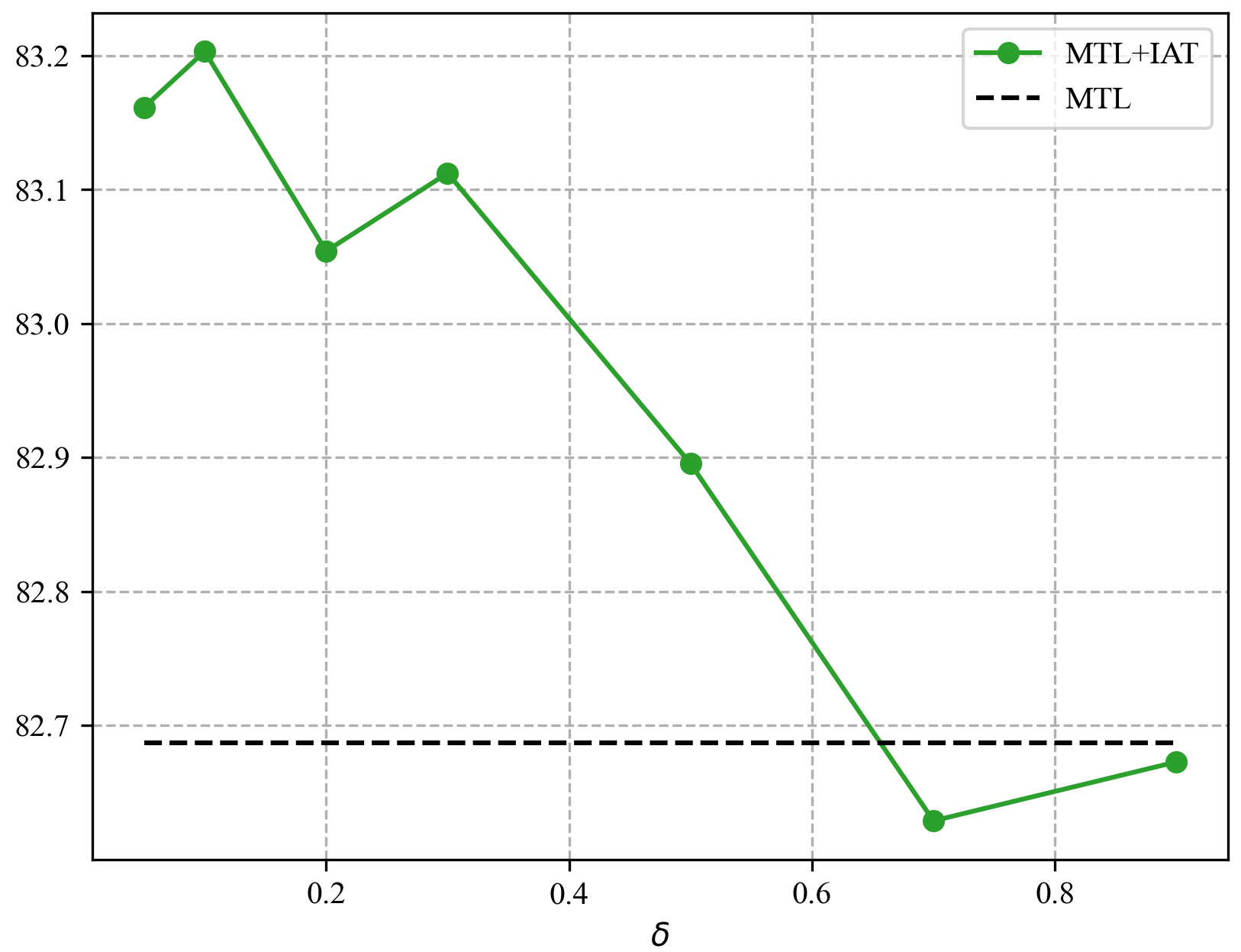}
\caption{The average performance of $\text{BERT}_{\text{BASE}}$ on GLUE dev set using IAT with different $\delta$ ($\alpha=50\%$). 
}\label{fig:glue_dev_delta}
\vspace{-3mm}
\end{figure}

\begin{figure}[ht]
\centering 
\setlength{\textfloatsep}{2pt}
\setlength{\intextsep}{2pt}
\setlength{\abovecaptionskip}{2pt}
\includegraphics[width=0.45\textwidth]{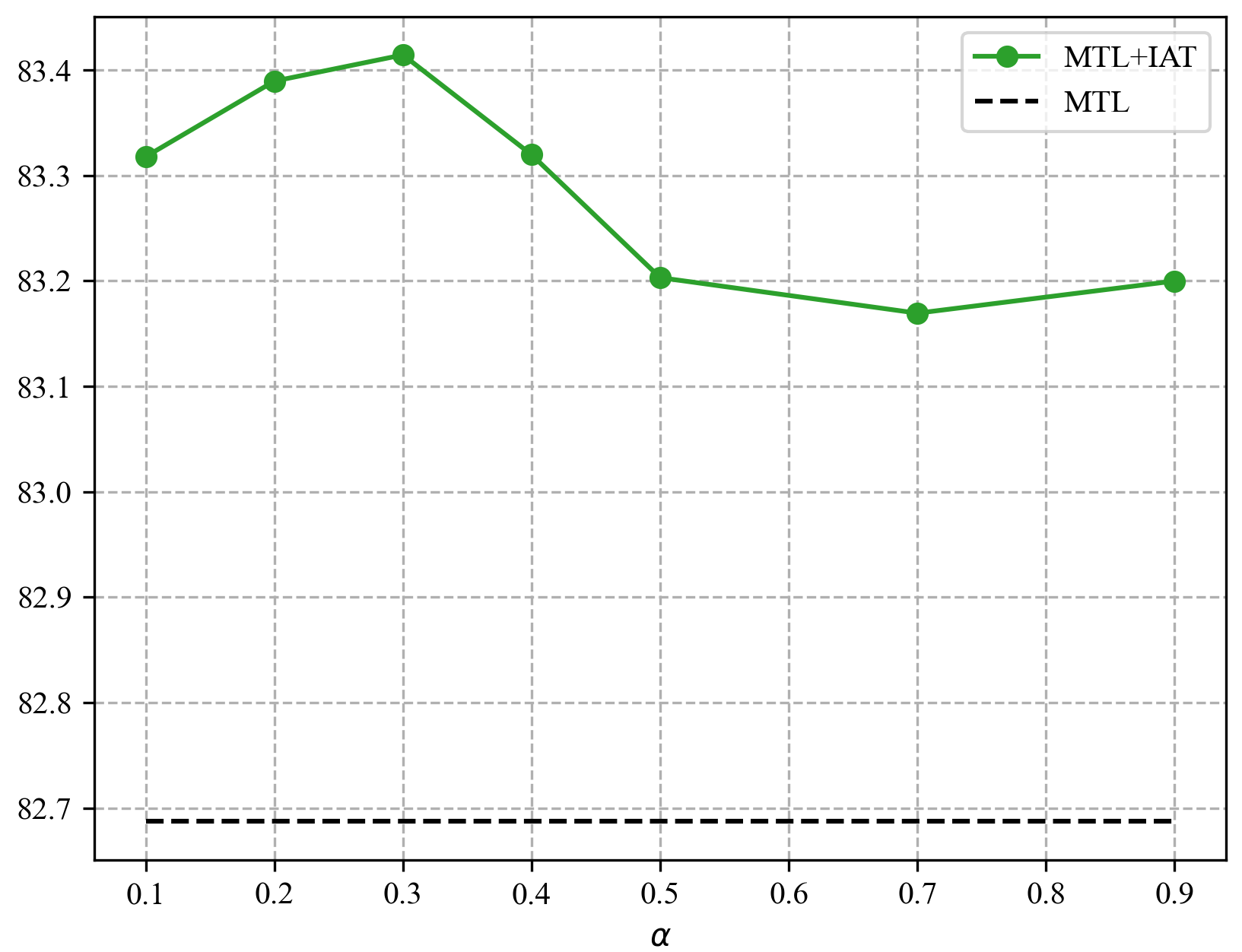}
\caption{The average performance of $\text{BERT}_{\text{BASE}}$ on GLUE dev set using IAT with different $\alpha$ ($\delta=10\%$). 
}\label{fig:glue_dev_theta}
\vspace{-3mm}
\end{figure}

Figure \ref{fig:glue_dev_delta} and \ref{fig:glue_dev_theta} present the impact of two hyperparameters, $\delta$ and $\alpha$ in IAT, on the average performance of $\text{BERT}_{\text{BASE}}$.
It is interesting to find that with a small $\delta$ and $\alpha$ (e.g., $\delta=10\%$ and $\alpha=30\%$), $\text{BERT}_{\text{BASE}}$ using IAT can achieve a good performance on GLUE dev set. 
Therefore, we only consider a limited hyperparameter sweep for each multi-task learning model with $\delta\ \in\ \{0.05, 0.1, 0.15\}$ and $\alpha\ \in\ \{0.1, 0.2, 0.3\}$.

Same as the finding in \citet{stickland2019pals}, the annealed sampling method is better for multi-task learning of GLUE than the proportional sampling method. 
The sampling probabilities of task $i$ in annealed sampling are changed with epoch $e$, and are calculated as follows:
\begin{equation}
    \label{eq:annealed}
    \small
    \begin{split}
        p_i &\propto N_i^{\varepsilon} \\
        \text{with}&\ \varepsilon=1-0.8\frac{e-1}{E-1}
    \end{split}
\end{equation}
where $N_i$ is the number of samples in task $i$, $E$ is the total number of epochs. 
In contrast, the $\varepsilon$ in proportional sampling is always equal to 1.

Table \ref{tab:glue_test_proportional} shows the results of five multi-task learning models using the proportional sampling method on GLUE test set. 
We can find that these multi-task learning models with proportional sampling perform better on GLUE test set after using IAT (+0.68\% on average), which is in line with the findings in Section \ref{sec:improve_glue}. 
It further demonstrates the effectiveness of our method.

Additional experimental results on development sets of GLUE for all models tested in this paper are reported in Table \ref{tab:glue_dev}. 
In most cases, the standard deviation of average performance on GLUE development set is less than or equal to the baseline after using IAT, which indicates the robustness of our method. 

\begin{table*}[htp]

\renewcommand\arraystretch{1.2}

\centering
\small

\setlength{\tabcolsep}{1.2mm}
 \begin{tabu}{rcccccccccc}
 
 \toprule[1.2pt]
\multicolumn{1}{c}{\textbf{Model}} & \begin{tabular}{@{}c@{}}\textbf{\small CoLA} \\ \scriptsize Mcc\end{tabular} & \begin{tabular}{@{}c@{}}\textbf{\small MNLI-(m/mm)} \\ \scriptsize Acc\end{tabular} & \begin{tabular}{@{}c@{}}\textbf{\small MRPC} \\ \scriptsize F1\end{tabular} & \begin{tabular}{@{}c@{}}\textbf{\small QNLI} \\ \scriptsize Acc\end{tabular} & \begin{tabular}{@{}c@{}}\textbf{\small QQP} \\ \scriptsize F1\end{tabular} & \begin{tabular}{@{}c@{}}\textbf{\small RTE} \\ \scriptsize Acc\end{tabular} & \begin{tabular}{@{}c@{}}\textbf{\small SST-2} \\ \scriptsize Acc\end{tabular} & \begin{tabular}{@{}c@{}}\textbf{\small STS-B} \\ \scriptsize $r^{s}$\end{tabular}& \textbf{\small Avg.}\\
   \midrule[0.8pt]
 $\text{TinyBERT}$  & $27.0$          & $\textbf{82.8}/\textbf{82.5}$& $83.4$         &$90.3$           & $\textbf{70.4}$ &  $71.6$          & $92.3$           & $83.4$            & $76.0$ \\ 
 \specialrule{0em}{0pt}{0pt}

 \ \ \ \ +IAT       & $\textbf{33.7}$ & $\textbf{82.8}/82.2$        & $\textbf{85.3}$ & $\textbf{90.6}$ & $70.3$          &  $\textbf{71.9}$ & $\textbf{92.5}$  & $\textbf{84.0}$   & $\textbf{77.0}$ \\ 
 \specialrule{0em}{0pt}{0pt}

 $\text{BERT}_{\text{BASE}}$    & $41.8$          & $83.6/82.7$                   & $85.0$          &  $\textbf{90.1}$  & $70.6$          &  $74.7$          & $\textbf{93.0}$ & $83.2$    & $78.3$ \\ 
 \specialrule{0em}{0pt}{0pt}

 \ \ \ \ +IAT                   & $\textbf{45.1}$ & $\textbf{84.0}/\textbf{83.3}$ & $\textbf{85.1}$ &  $89.6$           & $\textbf{70.8}$ &  $\textbf{76.2}$ & $92.8$          & $\textbf{83.5}$             & $\textbf{78.9}$ \\ 
 \specialrule{0em}{0pt}{0pt}
   
    % \midrule[0.8pt]
    
  $\text{BERT}_{\text{LARGE}}$  & $53.3$          & $85.4/\textbf{84.9}$            & $\textbf{85.9}$    &  $92.0$              & $71.3$          &  $78.6$          & $94.3$            & $84.8$    & $81.2$ \\ 
 \specialrule{0em}{0pt}{0pt}

 \ \ \ \ +IAT                   & $\textbf{56.7}$ & $\textbf{85.8}/84.8$            & $\textbf{85.9}$             &  $\textbf{92.3}$     & $\textbf{71.5}$ &  $\textbf{78.9}$ & $\textbf{94.4}$   & $\textbf{84.9}$             & $\textbf{81.7}$ \\ 
 \specialrule{0em}{0pt}{0pt}

    \midrule[0.8pt]
    
  $\text{RoBERTa}_{\text{BASE}}$  & $52.5$          & $\textbf{87.5}/\textbf{86.8}$  & $88.4$             &  $92.3$              & $71.9$          &  $79.1$          & $94.8$          & $\textbf{85.8}$    & $82.1$ \\ 
 \specialrule{0em}{0pt}{0pt}

 \ \ \ \ +IAT                     & $\textbf{56.1}$ & $87.1/86.7$                    & $\textbf{88.5}$    &  $\textbf{92.6}$     & $\textbf{72.3}$ &  $\textbf{79.9}$ & $\textbf{95.2}$   & $\textbf{85.8}$           & $\textbf{82.7}$ \\ 
 \specialrule{0em}{0pt}{0pt}
 
    \midrule[0.8pt]
    
   $\text{DeBERTa}_{\text{BASE}}$ & $61.9$          & $\textbf{89.9}/88.9$          & $87.6$             &  $93.7$              & $73.8$          &  $86.0$          & $95.8$          & $88.4$              & $85.1$ \\ 
 \specialrule{0em}{0pt}{0pt}

 \ \ \ \ +IAT                     & $\textbf{64.8}$ & $89.7/\textbf{89.0}$           & $\textbf{89.2}$    &  $\textbf{93.8}$     & $\textbf{73.9}$ &  $\textbf{86.7}$ & $\textbf{96.1}$   & $\textbf{88.8}$   & $\textbf{85.8}$ \\ 
 \specialrule{0em}{0pt}{0pt}
 
\bottomrule[1.2pt]
\end{tabu}
\caption{\label{tab:glue_test_proportional} GLUE test set results of five multi-task learning models using proportional sampling.}
\end{table*}

\begin{table*}[htp]

\renewcommand\arraystretch{1.2}

\centering
\small

\setlength{\tabcolsep}{1.2mm}
 \begin{tabu}{rccccccccc}
 
 \toprule[1.2pt]
\multicolumn{1}{c}{\textbf{Model}} & \begin{tabular}{@{}c@{}}\textbf{\small CoLA} \\ \scriptsize Mcc\end{tabular} & \begin{tabular}{@{}c@{}}\textbf{\small MNLI-(m/mm)} \\ \scriptsize Acc\end{tabular} & \begin{tabular}{@{}c@{}}\textbf{\small MRPC} \\ \scriptsize F1\end{tabular} & \begin{tabular}{@{}c@{}}\textbf{\small QNLI} \\ \scriptsize Acc\end{tabular} & \begin{tabular}{@{}c@{}}\textbf{\small QQP} \\ \scriptsize F1\end{tabular} & \begin{tabular}{@{}c@{}}\textbf{\small RTE} \\ \scriptsize Acc\end{tabular} & \begin{tabular}{@{}c@{}}\textbf{\small SST-2} \\ \scriptsize Acc\end{tabular} & \begin{tabular}{@{}c@{}}\textbf{\small STS-B} \\ \scriptsize $r^{s}$\end{tabular}& \textbf{\small Avg.}\\
\midrule[0.8pt]
\multicolumn{10}{c}{\textit{Proportional Sampling}} \\
\midrule[0.8pt]
  $\text{TinyBERT}$ & $31.3$         & $\textbf{83.2}/\textbf{83.5}$& $83.9$          &  $\textbf{90.3}$  & $86.3$          &  $71.8$          & $91.5$         & $85.8$          & $78.6_{\pm 0.5}$ \\ 
 \specialrule{0em}{0pt}{0pt}

 \ \ \ \ +IAT       & $\textbf{37.5}$ & $83.1/83.1$                 & $\textbf{85.5}$ &  $90.5$           & $\textbf{86.7}$ &  $\textbf{72.1}$ & $\textbf{91.6}$          & $\textbf{86.6}$ & $\textbf{79.6}_{\pm 0.3}$ \\ 
 \specialrule{0em}{0pt}{0pt}

  $\text{BERT}_{\text{BASE}}$     & $47.2$         & $\textbf{83.7}/83.3$ & $85.0$          &  $\textbf{90.3}$  & $87.0$          &  $78.3$          & $\textbf{92.7}$ & $86.6$          & $81.6_{\pm 0.3}$ \\ 
 \specialrule{0em}{0pt}{0pt}

 \ \ \ \ +IAT    & $\textbf{52.0}$ & $83.6/\textbf{83.6}$ & $\textbf{86.5}$ &  $90.1$           & $\textbf{87.2}$ &  $\textbf{79.4}$ & $91.9$          & $\textbf{87.2}$ & $\textbf{82.4}_{\pm 0.2}$ \\ 
 \specialrule{0em}{0pt}{0pt}
 
  $\text{BERT}_{\text{LARGE}}$     & $54.1$         & $85.7/85.6$                     & $\textbf{86.9}$   &  $91.6$           & $\textbf{88.3}$   &  $82.1$          & $93.2$         & $\textbf{87.9}$ & $83.9_{\pm 0.3}$ \\ 
 \specialrule{0em}{0pt}{0pt}

 \ \ \ \ +IAT    & $\textbf{58.9}$ & $\textbf{86.0}/\textbf{85.7}$  & $\textbf{86.9}$            &  $\textbf{91.8}$  & $88.2$            &  $\textbf{82.7}$ & $\textbf{93.4}$& $\textbf{87.9}$ & $\textbf{84.6}_{\pm 0.3}$ \\ 
 \specialrule{0em}{0pt}{0pt}
 
   \midrule[0.8pt]
   
  $\text{RoBERTa}_{\text{BASE}}$     & $49.7$         & $\textbf{87.6}/\textbf{87.1}$ & $89.7$  &  $91.9$             & $87.6$          &  $83.0$          & $\textbf{94.5}$ & $\textbf{88.4}$  & $84.4_{\pm 0.3}$ \\ 
 \specialrule{0em}{0pt}{0pt}

 \ \ \ \ +IAT    & $\textbf{54.2}$    & $87.2/\textbf{87.1}$ & $\textbf{89.8}$           &  $\textbf{92.3}$    & $\textbf{87.8}$ &  $\textbf{84.4}$ & $94.1$          & $\textbf{88.4}$           & $\textbf{85.0}_{\pm 0.1}$ \\ 
 \midrule[0.8pt]

  $\text{DeBERTa}_{\text{BASE}}$     & $65.5$         & $\textbf{89.8}/\textbf{90.0}$ & $89.1$            &  $93.8$             & $89.3$          &  $87.0$          & $95.1$         & $90.0$         & $87.7_{\pm 0.2}$ \\ 
 \specialrule{0em}{0pt}{0pt}

 \ \ \ \ +IAT    & $\textbf{67.9}$  & $89.7/\textbf{90.0}$          & $\textbf{90.1}$   &  $\textbf{93.9}$    & $\textbf{89.4}$ &  $\textbf{87.6}$ & $\textbf{95.5}$& $\textbf{90.1}$ & $\textbf{88.2}_{\pm 0.1}$ \\ 
 \midrule[0.8pt]
 
\multicolumn{10}{c}{\textit{Annealed Sampling}} \\
\midrule[0.8pt]
  $\text{TinyBERT}$ & $40.7$         & $\textbf{83.1}/82.9$  & $85.0$          &  $\textbf{90.4}$  & $86.2$          &  $73.6$          & $90.6$         & $87.5$          & $80.0_{\pm 0.3}$ \\ 
 \specialrule{0em}{0pt}{0pt}

 \ \ \ \ +IAT       & $\textbf{45.8}$ & $82.8/\textbf{83.0}$ & $\textbf{85.5}$ &  $90.3$           & $\textbf{86.6}$ &  $\textbf{74.7}$ & $\textbf{91.2}$& $\textbf{87.9}$ & $\textbf{80.9}_{\pm 0.4}$ \\ 
 \specialrule{0em}{0pt}{0pt}

  $\text{BERT}_{\text{BASE}}$     & $51.1$         & $83.6/83.9$                    & $87.6$          &  $90.1$         & $87.1$          &  $79.8$          & $92.2$           & $\textbf{88.4}$          & $82.6_{\pm 0.1}$ \\ 
 \specialrule{0em}{0pt}{0pt}

 \ \ \ \ +IAT                     & $\textbf{53.5}$ & $\textbf{83.8}/\textbf{84.0}$ & $\textbf{89.0}$ &  $\textbf{90.6}$& $\textbf{87.6}$ &  $\textbf{80.6}$ & $\textbf{93.2}$  & $\textbf{88.4}$ & $\textbf{83.4}_{\pm 0.2}$ \\ 
 \specialrule{0em}{0pt}{0pt}

  $\text{BERT}_{\text{LARGE}}$     & $58.8$         & $85.8/85.8$                     & $87.4$          &  $\textbf{91.8}$  & $87.9$   &  $82.0$          & $92.8$         & $88.6$          & $84.5_{\pm 0.2}$ \\ 
 \specialrule{0em}{0pt}{0pt}

 \ \ \ \ +IAT                      & $\textbf{61.6}$ & $\textbf{86.0}/\textbf{86.0}$  & $\textbf{88.7}$ &  $91.5$           & $\textbf{88.0}$   &  $\textbf{82.2}$ & $\textbf{93.7}$& $\textbf{89.1}$ & $\textbf{85.2}_{\pm 0.2}$ \\ 
 \specialrule{0em}{0pt}{0pt}
 
\midrule[0.8pt]

   $\text{RoBERTa}_{\text{BASE}}$     & $54.3$        & $\textbf{87.3}/\textbf{87.0}$ & $\textbf{92.2}$ &  $\textbf{92.3}$  & $86.9$          &  $84.6$  & $\textbf{94.5}$     & $89.0$          & $85.3_{\pm 0.1}$ \\ 
 \specialrule{0em}{0pt}{0pt}

 \ \ \ \ +IAT                     & $\textbf{59.2}$  & $87.1/86.8$                    & $\textbf{92.2}$  &  $92.2$          & $\textbf{87.1}$ &  $\textbf{84.7}$           & $94.3$              & $\textbf{89.1}$ & $\textbf{85.9}_{\pm 0.1}$ \\ 
 \specialrule{0em}{0pt}{0pt}
\midrule[0.8pt]

    $\text{DeBERTa}_{\text{BASE}}$     & $65.7$        & $\textbf{90.0}/\textbf{90.1}$ & $90.7$          &  $93.9$           & $89.0$          &  $88.0$            & $95.2$     & $90.3$          & $88.1_{\pm 0.2}$ \\ 
 \specialrule{0em}{0pt}{0pt}

 \ \ \ \ +IAT                     & $\textbf{68.7}$  & $89.9/90.0$                    & $\textbf{91.4}$  &  $\textbf{94.0}$  & $\textbf{89.1}$ &  $\textbf{88.5}$   & $\textbf{95.3}$              & $\textbf{90.7}$ & $\textbf{88.6}_{\pm 0.1}$ \\ 
 \specialrule{0em}{0pt}{0pt}
 
\bottomrule[1.2pt]
\end{tabu}
\caption{\label{tab:glue_dev} GLUE development set results of five multi-task learning models.
}
\end{table*}

\end{document}